%% file: main_arxiv.tex
\documentclass{article}
\usepackage[T1]{fontenc}
\usepackage{xcolor}
\usepackage{graphicx}
\usepackage{array}
\usepackage{booktabs}
\usepackage{multirow}
\usepackage{iclr2027_conference,times}
\usepackage{arxiv_preprint}

\input{math_commands.tex}

\usepackage{hyperref}
\usepackage{url}
\usepackage{enumitem}
\usepackage{listings}
\setlist[itemize]{leftmargin=*}

\definecolor{promptbackground}{rgb}{0.95,0.95,0.92}
\lstdefinestyle{prompt}{
  basicstyle=\ttfamily\scriptsize,
  backgroundcolor=\color{promptbackground},
  frame=single,
  rulecolor=\color{black!35},
  framerule=0.4pt,
  framesep=5pt,
  breaklines=true,
  breakatwhitespace=false,
  columns=fullflexible,
  keepspaces=true,
  showstringspaces=false,
  upquote=true,
  aboveskip=0.8\baselineskip,
  belowskip=0.8\baselineskip,
  literate={—}{{\textemdash}}1
}
\newcommand{\benchmarkname}{MotionGraphicsBench}
\newcommand{\numInstances}{100}
\newcommand{\writer}{\emph{verification program writer}}
\newcommand{\synthesizer}{\emph{test case synthesizer}}
\title{\benchmarkname{}:\\Evaluating Motion Graphics Generation\\with Programmatic Verification}

\author{
  Jiaju Ma\textsuperscript{1,2}\thanks{These authors contributed equally.} \quad
  Ashvini Kumar Jindal\textsuperscript{2}\footnotemark[1] \quad
  Chuhua Wang\textsuperscript{2} \quad
  Sumithra Bhakthavatsalam\textsuperscript{2} \\
  \normalfont\textsuperscript{1}Stanford University \qquad
  \textsuperscript{2}Figma
}
\date{}

\begin{document}

\maketitle

\begin{abstract}
\input{abstract}
\end{abstract}

\input{intro}

\input{related_work}
\input{benchmark_input}
\input{verifier_construction}

\input{experiments}

\input{results}
\input{conclusion}

\subsubsection*{Acknowledgments}
We thank Maneesh Agrawala for valuable discussions.
We thank our annotators for their careful judgments and Chinmaya Vempati for coordinating the annotation process.

\bibliography{references}
\bibliographystyle{references}

\appendix

\input{appendix_benchmark_input}

\input{appendix_verifier_construction}

\input{appendix_experiments}

\input{appendix_results}

\end{document}

%% file: math_commands.tex
\usepackage{amsmath,amsfonts,bm}

\def\eqref#1{equation~\ref{#1}}

\def\1{\bm{1}}

\DeclareMathAlphabet{\mathsfit}{\encodingdefault}{\sfdefault}{m}{sl}
\SetMathAlphabet{\mathsfit}{bold}{\encodingdefault}{\sfdefault}{bx}{n}



%% file: abstract.tex
%

Large language models can generate motion graphics from static images and natural language requests, but it remains difficult to evaluate how well models can follow requests that combine spatial, appearance, and temporal requirements.
We introduce \benchmarkname{}, an expert-curated benchmark of \numInstances{} instances for animating an input static SVG with a motion edit prompt.
By supplying all required visual elements, our editing task reflects a designer-like workflow and isolates motion generation from graphic generation.
%
Derived from real-world Lottie motion designs, our benchmark contains diverse combinations of spatio-temporal animation behaviors.
%
Each benchmark instance pairs its model input with a deterministic, human-readable verification program that accepts different valid implementations and reports which prompt requirements pass or fail, rather than comparing against a reference animation or using a VLM judge.
We construct these programs with a multi-agent pipeline using human-labeled animations and synthesized test cases.
%
On a balanced test set of 180 animations, our programs achieve 93.3\% raw agreement with human judgments, outperforming frontier VLM judges by 4.4\%--12.8\% while achieving at least 90\% recall for both human approvals and rejections.
Across 47 configurations from 32 LLMs and 12 providers, pass@1 ranges from 5.0\% to 68.0\%.
Across 15 models at their highest reasoning effort, additional attempts improve every model, but pass@32 still ranges from 38.0\% to 95.0\%.
%
Our analysis of verification programs reveals that generated animations often fail to match reference poses, follow requested motion trajectories, or coordinate related events.
%
Further, we fine-tune Qwen3.8-27B with group relative policy optimization (GRPO) using our verification programs as rewards, improving pass@1 from 30.31\% to 52.19\% on held-out instances.
Our project page is at \url{https://motion-graphics-bench.github.io/}.

%% file: intro.tex
\section{Introduction}
\label{sec:intro}

Motion graphics animate static imagery through coordinated changes in spatial structure, appearance and visibility, and temporal sequencing.
Large language models (LLMs) can now generate such animations from static images and natural language prompts \citep{tseng2025keyframer,park2025decomate}, but it remains difficult to evaluate how well the resulting animation realizes the requested motion: a single user request may combine multiple spatio-temporal conditions, but an animation may satisfy only some of it while missing others.
Natural language motion requests often describe salient animation behaviors and relative spatial and temporal relationships rather than exact parameters \citep{talmy1975motion,talmy1983language,tversky1998space}.
The same input can therefore correspond to several valid animations that differ in their trajectories and timing \citep{Ma2026selfconsistency}.
%
Existing related benchmarks on animation and video often evaluate outputs through metrics that compute resemblance to a target ``ground truth'' \citep{chen2026lottiegpt,yang2026omnilottie,huang2024vbench,ling2025vmbench}, or through VLM-based judges that assess rendered frames against the prompt \citep{he2024videoscore,lin2025vcode}.
Comparison with a single target is too restrictive because a generated animation can satisfy every requested condition while differing from the reference in its trajectory or timing.
VLM-based judges avoid comparison with a single reference animation.
However, their judgments can be stochastic and it is difficult to inspect how individual prompt requirements affect their decisions.

We introduce \benchmarkname{}, an expert-curated benchmark of \numInstances{} instances for evaluating vector animation generation.
We formulate motion graphics generation as an editing task: given an input static SVG and a motion edit prompt in natural language, a model generates an animation by adding requested motions to the static image, which contains all visual elements needed for the edit.
This setup mirrors a designer-like workflow in which the static design is first created before motions are added, and isolates motion generation from the separate problem of generating the graphic itself.

To ground our benchmark in real world motion designs rather than synthesized examples, we derive our instances from LottieAnimation-660K, a large collection gathered from professional animation websites \citep{chen2026lottiegpt}.
%
To more comprehensively evaluate model abilities, we curate our benchmark instances with diverse combinations of spatial, temporal, and appearance changes that challenge LLMs while remaining clearly describable in natural language.
%
Our final benchmark covers 17 animation requirement types grouped into spatial structure, appearance and visibility, and temporal sequencing.
%
We rank source animations by structural complexity and manually retain moderately complex, unique motions that can be clearly described in natural language.
%
For each selected source animation, we construct an input static SVG and motion edit prompt through LLM assistance and human review.
The SVG provides a reference state for the animated elements while retaining every visual element needed to realize the edit.
Our motion edit prompts request salient animation behaviors and their relative spatial and temporal relationships.
The prompt states what should move or change, how elements should relate, and when actions should occur without prescribing exact measurements \citep{talmy1975motion,talmy1983language,tversky1998space,Ma2025mover,Ma2026selfconsistency}.

%
%

\begin{figure}[t]
    \setlength{\abovecaptionskip}{4pt}
    \centering
    \includegraphics[width=\linewidth]{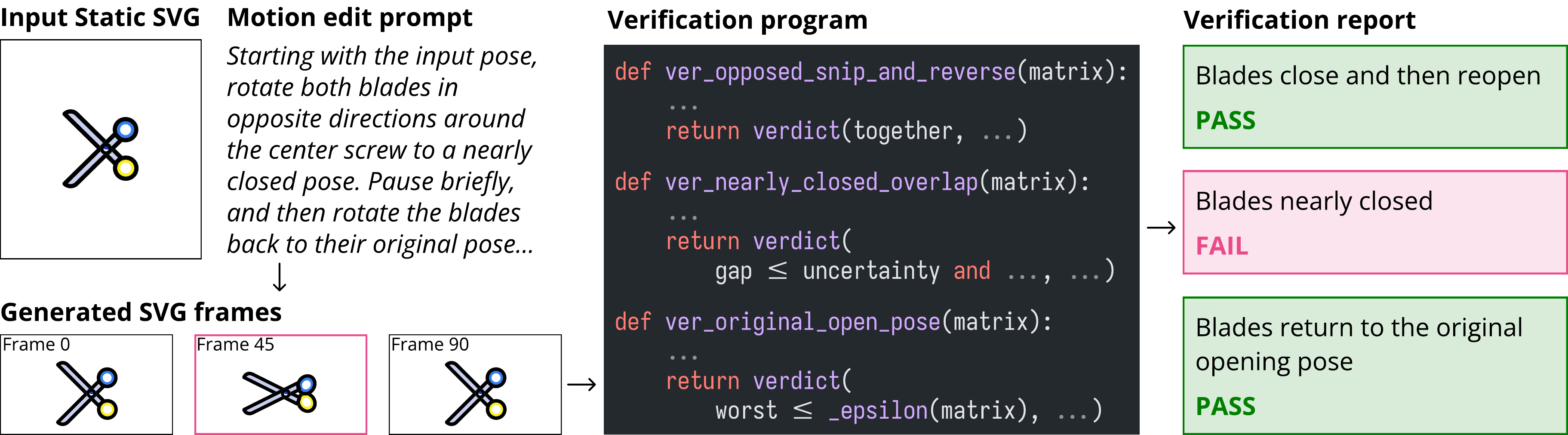}
    \caption{\textbf{Overview of \benchmarkname{}.}
    Given an input static SVG and a motion edit prompt, our benchmark uses a human-readable verification program to evaluate whether the generated animation satisfies the input requirements and outputs a report. Here, the scissors have the correct motion sequence but fail the verification because the blades remain too far apart when closed.}
    \label{fig:overview}
    \vspace{-5mm}
\end{figure}

Because each motion edit prompt specifies different requirements, evaluating model-generated animations requires instance-specific checks.
Our deterministic evaluation framework therefore pairs each benchmark instance's model input with a human-readable verification program.
Each program consists of deterministic checks corresponding to the requirements in its motion edit prompt.
The program measures how visual elements change across the generated SVG frames and returns a verification report with the overall decision and individual check outcomes.

To construct one prompt-specific verification program for every benchmark instance, we combine human-labeled generated animations with a multi-agent construction pipeline.
%
To support both verification program construction and evaluation, we collect human judgments of whether model-generated animations satisfy their motion edit prompts.
We use human-labeled construction animations to build the programs and reserve a held-out test set for evaluating the resulting programs.
%
Our pipeline consists of a verification program writer and a test case synthesizer.
The writer drafts and revises the program using the motion edit prompt, source Lottie animation, human-labeled construction animations, and synthesized test cases, while the synthesizer creates and reviews test cases.

Deterministic decisions are useful only if the verification programs capture human judgments of whether an animation satisfies the prompt.
We therefore evaluate our programs on 180 human-labeled test animations.
Our programs achieve 93.3\% raw agreement, 90.0\% approval recall, and 96.7\% rejection recall.
As a comparison, we use four frontier VLMs as judges and they only achieve 80.6\%--88.9\% raw agreement on the same animations, outperformed by our verification programs by 4.4\%--12.8\%.
We observe that the VLM judges are either overly lenient or have both approval and rejection recall below 85\%.

Using our verification programs, we evaluate 47 configurations from 32 LLMs across 12 providers on all \numInstances{} benchmark instances.
Across these configurations, pass@1 ranges from 5.0\% to 68.0\%.
%
From the 47 configurations, we select the highest reasoning effort configuration for 15 models and run 32 independent generation attempts per benchmark instance.
Additional attempts improve model performances, but pass@32 still ranges from 38.0\% to 95.0\%.
We fine-tune Qwen3.8-27B with group relative policy optimization (GRPO) using verification program rewards, improving pass@1 from 30.31\% to 52.19\% and pass@32 from 60.0\% to 100.0\% on ten instances held out from GRPO training.
%
By examining our verification programs and reports, we identify three main failure modes across our benchmark.
Generated animations may fail to match reference poses, follow requested motion trajectories, or coordinate related events.
%
Together, \benchmarkname{} provides a curated benchmark grounded in real world motion designs, deterministic evaluation through interpretable verification programs matching human judgments, and a broad analysis of model performance and generated animation failures.
Our project page is at \url{https://motion-graphics-bench.github.io/}.

%% file: related_work.tex
\section{Related Work}
\label{sec:related_work}

\paragraph{Vector animation generation.}
Recent vector animation systems either use end-to-end generation or editing setups.
LottieGPT and OmniLottie generate both graphics and motion from text or visual conditions, so their evaluations reflect both capabilities \citep{chen2026lottiegpt,yang2026omnilottie}.
Editing methods begin with existing graphics: Keyframer and Decomate support language-guided SVG editing, AniClipart optimizes clipart motion, Vector Prism, LiveSVG, and VAnim animate supplied vector graphics, and MoVer adds and verifies prompt-specified motion in existing SVG scenes \citep{tseng2025keyframer,park2025decomate,wu2024aniclipart,yun2026vectorprism,levy2026livesvg,liang2026vanim,Ma2025mover}.
Our benchmark follows the editing setup: an input static SVG supplies every required visual element, and a motion edit prompt derived from a source Lottie animation specifies the motion, isolating motion generation from graphic generation.

\paragraph{Benchmark task construction.}
Existing benchmarks either construct prompts around predefined capabilities or curate tasks from existing artifacts or user requests.
VBench designs prompts by evaluation dimension and content category, while GenEval, ConceptMix, and T2V-CompBench target specific objects, attributes, relations, motions, or combinations \citep{huang2024vbench,ghosh2023geneval,liang2024conceptmix,sun2025t2vcompbench}.
MoVer and Self-Consistency similarly use synthetic template-based prompts to evaluate motion relations or trajectory families \citep{Ma2025mover,Ma2026selfconsistency}.
DS-1000 derives problems from Stack Overflow, SWE-bench from repository issues and tests, ScienceAgentBench from scientific datasets and tasks, and WildBench from real-user queries \citep{lai2023ds1000,jimenez2024swebench,chen2025scienceagentbench,lin2025wildbench}.
Our benchmark similarly selects existing Lottie animations and constructs motion edit prompts from their motion designs.
Our motion edit prompts span 17 animation requirement types, while each prompt-specific verification program evaluates one instance's exact requirements.

\paragraph{Verification of generated outputs.}
Because several animations can satisfy one motion edit prompt, evaluation should test its requested relationships without requiring a target trajectory.
VBench applies dimension-specific metrics to motion smoothness, temporal consistency, and spatial relationships, producing broad capability profiles rather than prompt-specific decisions over all requested behaviors \citep{huang2024vbench}.
Reference-based comparisons in LottieBench and MMLottie-Bench can penalize valid deviations, while VCode and VideoScore use flexible model-based judgments that depend on the judge and prompting procedure \citep{chen2026lottiegpt,yang2026omnilottie,lin2025vcode,he2024videoscore}.
%
%
%
MoVer maps prompts to first-order spatio-temporal predicates and evaluates its approach on 5,600 synthetic prompts generated from templates and LLM rewrites \citep{Ma2025mover}.
However, its predicates cannot verify requirements on easing, appearance changes, shape deformation, or the shape of a motion path \citep{Ma2025mover,Ma2026selfconsistency}, and at least one such requirement appears in 97 of our 100 instances (Appendix~\ref{app:requirement_types}).
To support verification of these requirements, we construct human-readable Python verification programs for each instance and validate their decisions against held-out human judgments.
%

%% file: benchmark_input.tex
%

\begin{table}[t]
    \centering
    \caption{\textbf{Benchmark coverage.}
    An animation requirement type identifies a spatial, appearance, or temporal condition specified by a motion edit prompt.
    The \emph{Instances} column counts benchmark instances containing at least one type in the family.
    See Appendix~\ref{app:requirement_types}.
    }
    \label{tab:animation_requirement_coverage}
    \small
    \setlength{\tabcolsep}{4pt}
    \renewcommand{\arraystretch}{1.08}
    \begin{tabular*}{\linewidth}{@{\extracolsep{\fill}}>{\raggedright\arraybackslash}p{0.21\linewidth}>{\centering\arraybackslash}p{0.06\linewidth}>{\centering\arraybackslash}p{0.09\linewidth}p{0.535\linewidth}@{}}
        \toprule
        Requirement family & Types & Instances & Highest-coverage types (instances) \\
        \midrule
        Spatial structure & 8 & 100 & \emph{Geometric state constraints} (99); \emph{part attachment and group integrity} (78); \emph{translation trajectories} (72) \\
        Temporal sequencing & 5 & 100 & \emph{Temporal ordering} (100); \emph{temporal coordination and phase} (93); \emph{animation speed and easing} (93) \\
        Appearance and visibility & 4 & 65 & \emph{Visibility and opacity} (53); \emph{occlusion and visible-region containment} (30); \emph{color, fill, and stroke appearance} (13) \\
        \bottomrule
    \end{tabular*}
    \vspace{-2mm}
\end{table}

\section{\benchmarkname{}}
\label{sec:benchmark}

\subsection{Task Formulation}
\label{sec:task_formulation}

\textbf{Model inputs and output.}
To isolate motion generation from graphic creation, we formulate vector animation generation as a motion editing task.
We provide each model with an \emph{input static SVG} and a \emph{motion edit prompt} written in natural language, then ask it to generate an animation that realizes the requested motion.
An \emph{input static SVG} is a vector scene that contains all scene geometry, including elements that may not be visible when rendered.
A \emph{motion edit prompt} describes how the input static SVG should be animated, including what should move or change, how elements should relate, and when actions should occur.
Because it is often difficult for users to specify exact keyframes, coordinates, or frame numbers (e.g., how many degrees the scissor blades in Figure~\ref{fig:overview} should rotate), each prompt in our benchmark requests salient animation behaviors in terms of relative spatial and temporal relationships, consistent with findings from cognitive psychology on how people describe animation \citep{talmy1975motion,talmy1983language,tversky1998space}.
As a result, a prompt often does not fully specify a unique animation \citep{Ma2026selfconsistency}.
The input static SVG provides a reference appearance, but it does not necessarily define frame 0.
The prompt may ask the model to construct a different frame 0 based on the input SVG or to make the animation reach the SVG's depicted state later (e.g., the \emph{Scissors} prompt asks the blades to return to the open pose shown in the input SVG in Figure~\ref{fig:overview}).

\textbf{Evaluating model outputs.}
Multiple model outputs may therefore satisfy the same prompt while differing in their trajectories and timing.
To evaluate a model output against animation requirements rather than one target animation, each \emph{benchmark instance} pairs an input static SVG and motion edit prompt with a prompt-specific verification program.
The program encodes the animation requirements as deterministic, human-readable checks.
Section~\ref{sec:evaluation_framework} describes how our evaluation framework applies the program to model-generated animations.

\subsection{Model Input Construction}
\label{sec:benchmark_input_construction}

\textbf{Source Lottie corpus selection.}
To ground our benchmark in real-world motion designs rather than synthetic examples, we select \numInstances{} source Lottie animations from LottieAnimation-660K, a collection gathered from professional animation websites \citep{chen2026lottiegpt}.
Lottie preserves layers, shapes, transforms, keyframes, and easing curves, which lets us identify the elements and motions needed to construct each model input.
We use each source Lottie animation only to construct the input static SVG and motion edit prompt.
To keep the task focused on motion generation rather than identifying unnamed SVG elements, we retain animations whose animated layers or objects have names that support clear references in a motion edit prompt.
We rank the remaining candidates by their animated property types, animated objects, and keyframes to capture motion variety, scene complexity, and temporal structure.
We manually review 1,000 candidates near the median complexity to balance challenge and describability, removing duplicate graphics and motions and animations that cannot be clearly described in natural language.

\textbf{Input static SVG construction.}
To model a workflow in which an artist supplies an existing static design, we exclude frames that are empty, cropped, or transitional.
We select a remaining frame in which the visible objects are recognizable, opaque, and normally sized.
To keep our benchmark focused on motion editing rather than graphic creation, the input static SVG must contain every visual element needed by the requested animation, including elements hidden in the selected frame.
A single-frame Lottie render can omit elements that are inactive or fully transparent at the selected moment, even when the requested animation needs them later.
We therefore recover this hidden geometry while preserving the selected frame's visible appearance \citep{liang2026vanim}.
Appendix Figure~\ref{fig:input_static_svg_construction} illustrates the process.

\textbf{Motion edit prompt construction.}
Because the source Lottie animations do not include model-facing instructions for adding motion to a static SVG, we construct each motion edit prompt through LLM-assisted drafting and human review.
As motivated in Section~\ref{sec:task_formulation}, each prompt states what should move or change, how elements should relate, and when actions should occur.
We describe these requirements through relative spatial and temporal relationships rather than exact keyframes, coordinates, or frame numbers \citep{Ma2025mover}.
When needed, a prompt specifies a visible path, timing pattern, or easing behavior without requiring exact reproduction of the source Lottie animation \citep{Ma2026selfconsistency}.
An LLM drafts each prompt from the input static SVG and evidence from the source Lottie animation.
The draft names the relevant elements and states their required spatial and temporal relationships, including end states.
These details distinguish a complete edit from one that captures only the general motion.
Two human reviewers participate in prompt review.
A reviewer compares each draft with the input static SVG and source Lottie animation, then approves it, edits it, or requests a targeted revision to remove ambiguity, unsupported claims, and implementation-specific wording.
All four authors additionally review 15 randomly sampled motion edit prompts.
This review procedure follows prior work that uses multiple reviewers during benchmark construction \citep{zhuo2025bigcodebench,chen2025scienceagentbench,lai2023ds1000}.
See Appendix~\ref{app:input_construction_protocol}.

\subsection{Benchmark Composition and Coverage}
\label{sec:benchmark_composition}

Our \benchmarkname{} contains \numInstances{} benchmark instances constructed from \numInstances{} distinct source Lottie animations.
Each instance pairs an input static SVG and a human-reviewed motion edit prompt with a prompt-specific verification program.
Our \numInstances{} instances are comparable in scale to Terminal-Bench 2.0 with 89 tasks, MatPlotBench with 100 visualization tasks, ScienceAgentBench with 102 scientific tasks, and Vector Prism with 114 SVG-instruction pairs \citep{merrill2026terminalbench,yang2024matplotagent,chen2025scienceagentbench,yun2026vectorprism}.
The source Lottie animations have a mean duration of 3.50 seconds (SD 1.89).
The motion edit prompts have a mean length of 193.8 words (SD 57.2).
To characterize the animation behaviors requested by our motion edit prompts, we identify 17 animation requirement types grouped into three families: spatial structure, appearance and visibility, and temporal sequencing (Table~\ref{tab:animation_requirement_coverage}).
Our benchmark centers on the temporal ordering and geometric state constraints needed to specify coherent motion, while extending its coverage to deformation, reveal and erasure, and color, fill, and stroke appearance.
Each motion edit prompt combines between 7 and 14 animation requirement types.
All prompts combine spatial and temporal requests; 65 include appearance or visibility.

%% file: verifier_construction.tex
\section{Evaluation with Verification Programs}
\label{sec:evaluation_framework}

\subsection{Verification of Generated Animations}
\label{sec:verification_interface}

\textbf{Prompt-specific verification.}
Motion edit prompts describe intended spatial and temporal behavior without fully specifying an exact trajectory, timing curve, or grouping \citep{talmy1975motion,talmy1983language,tversky1998space}.
Multiple animations can therefore satisfy the same prompt \citep{Ma2025mover,Ma2026selfconsistency}.
Because each prompt imposes different requirements, we evaluate each instance with a prompt-specific verification program rather than generic similarity measures.
The program implements the prompt requirements as verification checks that evaluate spatial structure, appearance and visibility, and temporal sequencing as applicable to the instance.
These checks allow different animation implementations to pass when they satisfy the requested behavior while detecting partial results that generic visual metrics may miss.
We describe how we construct a verification program for each instance in Section~\ref{sec:verifier_construction}.

\textbf{Verification procedure.}
Vector animations can be created through many mechanisms, such as Lottie, GSAP, anime.js, CSS animations, and SMIL \citep{chen2026lottiegpt,gsap,animeJs,cssAnimation,smil3}.
Animations produced through any of these mechanisms can be represented as sequences of SVG frames.
Because each frame records the vector scene at one point in time, the resulting sequence provides a common representation that allows verification to operate independently of how the motion is produced.
In our experiments, we ask models to generate JavaScript programs with GSAP, whose timeline API lets our renderer seek to fixed timestamps and deterministically capture the resulting SVG frames \citep{gsap,Ma2025mover}.
The verification program takes the SVG frames as input, measures each element's geometry, appearance, and motion over time, evaluates the prompt requirements, and returns a verification report with the overall decision, individual check outcomes, and any errors.
An attempt is one model request for one benchmark instance and counts as a success only if it produces an animation that passes every verification check.

\subsection{Verification Program Construction}
\label{sec:verifier_construction}

\begin{figure*}[t]
    \setlength{\abovecaptionskip}{4pt}
    \centering
    \includegraphics[width=\textwidth]{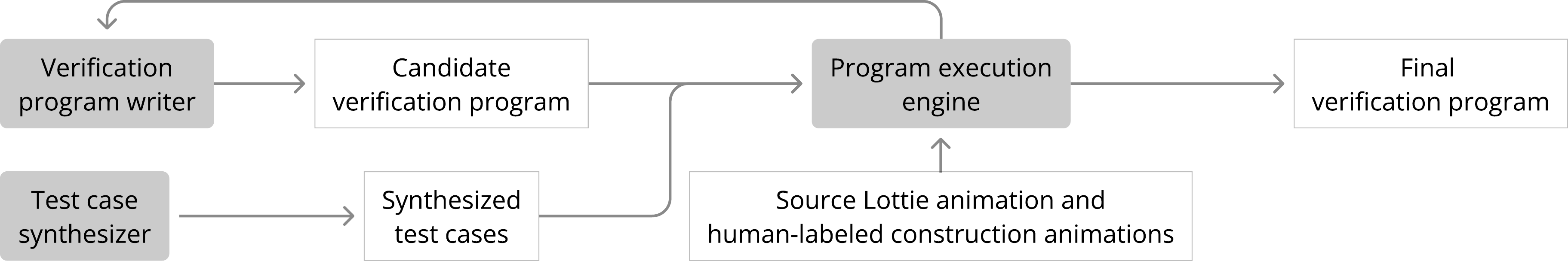}
    \caption{\textbf{Verification program construction.}
    The writer drafts and revises a candidate program using the benchmark inputs, source Lottie animation, and human-labeled construction animations, while the synthesizer independently creates test case animations.
    We run the candidate on these real and synthesized animations, and then return the reports to the writer for revision.}
    \label{fig:verification_program_construction}
    \vspace{-5mm}
\end{figure*}

Because a motion edit prompt can be satisfied by several valid implementations, each verification program must accept allowed variation while rejecting animation requirement violations.
We construct one program per instance with two agents: we have a \writer{} that writes and revises verification programs, while a \synthesizer{} creates synthesized test case animations that expose missing or overly restrictive checks (Figure~\ref{fig:verification_program_construction}).
Both agents use GPT-6 Astra.

\textbf{Construction data.}
To ground construction in diverse model outputs, we use a human-labeled construction set of 1,585 model-generated animations from 25 generation configurations across all \numInstances{} benchmark instances, comprising 410 approvals and 1,175 rejections.
Appendix Table~\ref{tab:appendix_model_settings} marks these 25 configurations and the four held-out configurations.
These animations come from our human judgment collection procedure described in Section~\ref{sec:human_labeled_evaluation}, where we ask human annotators to review model-generated animations against corresponding input static SVGs and motion edit prompts.
Our writer and synthesizer receive only construction animations, human review notes, instance inputs, and the source Lottie animation (Appendix~\ref{app:prompt_review}).
%

\textbf{Verification program writer.}
To keep each verification program grounded in the requested motions, the \writer{} receives a \emph{construction example set} alongside the input static SVG and motion edit prompt.
The set consists of the source Lottie animation, human-labeled construction animations, and synthesized test cases.
The source Lottie animation supplies one valid realization of the requested edit, while examples approved by humans show allowed variation and examples rejected by humans and their review notes identify possible violations to inspect.
Together, these inputs help the writer interpret the requested edit without treating one animation as the definition of correctness or adding requirements absent from the input static SVG or motion edit prompt.
The writer uses an \emph{animation matrix} of SVG attributes and measurements; Appendix~\ref{app:evaluation_implementation} gives implementation details.
The writer therefore bases its measurements and tolerances on the requested behavior and measured animation properties rather than choosing thresholds solely to match the construction approvals and rejections.
Using our \emph{verification library}, which derives spatio-temporal measurements  from the animation matrix, the writer implements these requirements as a Python verification program (Figure~\ref{fig:overview}).
During construction, the writer tests each candidate program on the complete construction example set.
The source Lottie animation and every approved or positive example must pass, while every rejected or negative example must fail.

\textbf{Test case synthesizer.}
Human-labeled construction animations and the source Lottie animation may leave a requirement untested or combine several violations in one example, causing a candidate verification program to become too permissive or overly restrictive.
The \synthesizer{} therefore creates synthesized test cases that probe allowed variation and targeted violations.
It receives the motion edit prompt, source Lottie animation, human-labeled construction animations and review notes, and descriptions of earlier cases, but not the candidate program or its reports.
Using our \emph{animation transformation library}, which deterministically changes motion timing, direction, extent, position, appearance, and SVG structure, it transforms the source Lottie animation or a construction animation approved by humans.
Positive cases preserve the requested edit while varying allowed choices, whereas negative cases introduce a violation.
When useful, paired cases isolate one relevant property.
To account for the imbalance between approvals and rejections, the synthesizer requests more negative cases when approvals are more common and more positive cases when rejections are more common.
The assigned labels remain claims made during construction rather than independent human judgments.
See Appendix~\ref{app:verifier_construction}.

\textbf{Testing and revision.}
After the writer produces a candidate verification program, it runs the program on the complete construction example set, inspects the individual check outcomes, revises the candidate when the tests expose missing or overly restrictive checks, and reruns the set.
A synthesized-label disagreement may reveal a faulty test case, which the agents resolve within the loop (Appendix~\ref{app:verifier_construction}).
This iterative construction loop resembles counterexample-guided synthesis and selection of generated tests \citep{solarlezama2008sketching,chen2023codet,jain2025testgeneval,he2026hardtestgen}.
The run produces one verification program per instance, and we freeze the resulting complete set for evaluation on the reserved test set (Section~\ref{sec:evaluation_validation}).

%% file: experiments.tex
\section{Verification Program Validation}
\label{sec:evaluation_validation}

\subsection{Measuring Alignment with Human Judgment}
\label{sec:human_labeled_evaluation}
\label{sec:validation_design}

\begin{table}[t]
\centering
\caption{
We compare our verification programs against four VLM judges as baseline on the same balanced test set of 180 animations.
Raw agreement is the fraction matching human labels; approval and rejection recall are the fractions of human approvals and rejections correctly identified.
}
\label{tab:validation_agreement}
\footnotesize
\begin{tabular*}{\textwidth}{@{\extracolsep{\fill}}lccc@{}}
\toprule
Method & Raw Agreement & Approval Recall & Rejection Recall \\
\midrule
Opus 5 & 80.6\% & \textbf{100.0\%} & 61.1\% \\
Opus 5.5 & 88.9\% & 96.7\% & 81.1\% \\
GPT-6 Astra & 83.3\% & 82.2\% & 84.4\% \\
GPT-6.1 Sol & 82.8\% & 83.3\% & 82.2\% \\
Our verification programs & \textbf{93.3\%} & 90.0\% & \textbf{96.7\%} \\
\bottomrule
\end{tabular*}
\vspace{-5mm}
\end{table}

\textbf{Human judgment collection.}
We collect human judgments for 1,980 model-generated animations, and use 1,585 of those for verification program construction (Section~\ref{sec:verifier_construction}) and reserve a disjoint set of 180 animations to evaluate how well the resulting verification programs align with human judgment.
Our annotation procedure has two phases: an independent annotation phase followed by a disagreement resolution phase, following established procedures used in prior works \citep{banea2008multilingualsubjectivityanal,hromei2025groundedsemanticrole,bansal2025videophy}.
In the independent annotation phase, two assigned annotators inspect the motion edit prompt, input static SVG, and generated animation and judge whether that animation satisfies the prompt rather than aesthetic preference or exact source animation reproduction.
Annotators may reveal the initially-hidden source Lottie animation to help them visualize what a motion edit prompt is asking for (Appendix Figure~\ref{fig:human_review_interface}).
When annotators reject an animation, we ask them to identify which parts of the motion edit prompt are not satisfied and explain why.
To keep judgments independent, we hide the generating model and allow each annotator to see only their own annotations.
Annotators do not inspect, write, or revise our verification programs and do not know which animations will be used for construction or held-out evaluation.
Our initial raw inter-annotator agreement is 88.88\%, with Krippendorff's $\alpha = 0.718$ and Fleiss' $\kappa = 0.718$.
For context, GenEval reports 87.8\% pairwise agreement, and Banea et al.\ report $\kappa=0.67$, rising to $0.77$ after excluding uncertain annotations \citep{ghosh2023geneval,banea2008multilingualsubjectivityanal}.
After all annotators finish their assigned independent judgments, we begin the disagreement resolution phase.
In this phase, the same pair revisits animations with conflicting decisions and discusses the prompt, and we assign a binary final label only when both annotators agree.
See Appendix~\ref{app:prompt_review} for more details.

\textbf{Held-out evaluation.}
To test whether our verification programs generalize to unseen animation distributions, we reserve outputs from Gemini 3.1 Pro, Gemma4, GPT-5.5, and Sonnet 5 for held-out evaluation and use outputs from other models for construction.
The four models used for held-out evaluation span multiple providers and include both open-weight and proprietary models.
Their outputs include 90 human-approved and 274 human-rejected animations.
To balance our test set, we discard 184 human-rejected animations, leaving 180 animations with 90 human approvals and 90 human rejections across all \numInstances{} benchmark instances.
We also evaluate our programs on all 364 animations and report the results in Appendix~\ref{app:complete_experiments}.
For comparison, we use Opus 5, Opus 5.5, GPT-6 Astra, and GPT-6.1 Sol as VLM judge baselines on the balanced test set.
Each judge receives the input static SVG rendering, motion edit prompt, and up to 64 timestamped frames from each generated animation.
See Appendix~\ref{app:complete_experiments} for the judge settings and prompt.

\textbf{Validation results.}
Our verification programs achieve 93.3\% raw agreement with human judgments on our balanced test set, outperforming the four VLM judge baselines by 4.4--12.8 percentage points (Table~\ref{tab:validation_agreement}).
We find that Opus 5 and Opus 5.5 are overly lenient, with approval recall of 100.0\% and 96.7\% but rejection recall of only 61.1\% and 81.1\%, respectively.
GPT-6 Astra and GPT-6.1 Sol have more balanced approval and rejection recall, but both remain relatively low at 82.2\%--84.4\%.
Our programs achieve 90.0\% approval recall and 96.7\% rejection recall.
Our programs also achieve higher raw agreement within each of the four generation configurations represented in the test set.
%

To understand the VLM judges' failure patterns, we examine all 116 decisions that differ from human judgments across 53 distinct model-generated animations.
%
To understand the errors in our verification programs' judgments, we review all 12 decisions that differ from human judgments on our balanced test set (see Appendix~\ref{app:complete_experiments} for details on both analyses).

\subsection{Ablation Studies of Verification Program Construction}
\label{sec:construction_components}

\textbf{Human-labeled examples and synthesized test cases.}
To understand how our construction examples help verification programs accept valid variations of animations that satisfy a motion edit prompt and reject violations, we compare three conditions in Table~\ref{tab:construction_components}.
Each condition constructs programs for all \numInstances{} instances using the same settings as our full construction procedure described in Section~\ref{sec:verifier_construction}, except for the ablated components.
To account for variation in LLM outputs, we report mean results across three runs per condition on the balanced test set of 180 animations.
%
Without human-labeled examples or synthesized test cases, our writer still receives the source Lottie animation as an example that satisfies the prompt.
The resulting programs tend to be overly strict, with 63.7\% approval recall compared with 92.6\% rejection recall.
Adding synthesized test cases improves approval recall to 73.3\%.
Further adding human-labeled examples improves both approval and rejection recall, reaching 90.0\% and 96.7\%, respectively.

\begin{table}[tb]
\centering
\caption{
Ablation of human-labeled examples and synthesized test cases on verification program performance.
We report means across three runs per condition on the balanced test set.
}
\label{tab:construction_components}
\resizebox{\textwidth}{!}{%
\begin{tabular}{@{}lccc@{}}
\toprule
Construction condition & Raw agreement & Approval recall & Rejection recall \\
\midrule
Without human-labeled examples or synthesized test cases & 78.2\% & 63.7\% & 92.6\% \\
Without human-labeled examples & 82.4\% & 73.3\% & 91.5\% \\
Full construction & \textbf{93.3\%} & \textbf{90.0\%} & \textbf{96.7\%} \\
\bottomrule
\end{tabular}%
}
\vspace{-5mm}
\end{table}

\textbf{Iterative revisions.}
To test whether iterative revisions between the writer and the synthesizer help our verification programs better match human judgments, we evaluate the initial programs and their revised versions in the Full condition.
We observe that mean agreement increases from 61.79\% for the initial programs to 93.33\% for the final programs, with the largest improvement after the first revision.
We report the construction settings and detailed results in Appendix~\ref{app:construction_studies}.

\subsection{Verification Programs for Reinforcement Learning}
\label{sec:reinforcement_learning}

To complement our held-out validation, we test whether our verification programs provide useful learning signals for improving animation generation.
Inspired by HardTestGen's use of test-based rewards and Flow-GRPO's use of GenEval-based rewards \citep{he2026hardtestgen,liu2025flowgrpo}, we fine-tune Qwen3.8-27B using group relative policy optimization (GRPO) on 90 benchmark instances and reserve the remaining ten for held-out RL evaluation.
The model receives the motion edit prompt, input static SVG, and timing constraints, but not the verification program code or source Lottie animation.
We reward each generated animation by the fraction of verification checks it passes and assign zero to invalid outputs.
We evaluate the base and trained models with 32 attempts per held-out instance.
After 41 policy updates, we observe pass@1 increasing from 30.31\% to 52.19\% and pass@32 from 60.0\% to 100.0\%.
Please see Appendix~\ref{app:reinforcement_learning} for the training details and full results.

%% file: results.tex
\section{Model Performance and Failure Analysis}
\label{sec:results}

We report pass@1 for 47 model configurations representing 32 LLMs from 12 providers, including 16 open-weight and 16 proprietary models.
We first generate one animation per benchmark instance for each configuration to evaluate pass@1.
For pass@k, we then select the highest-available-effort configuration from 15 models and collect 31 additional independent attempts per benchmark instance, bringing the total to 32.
This design yields 51,200 attempts across 100 instances.
All model scores use a separate generation batch from human-reviewed sets.
Our verification programs evaluate rendered SVG frames independently of the animation format.
For a controlled comparison, we ask every model to generate a JavaScript program with GSAP, which has been the industry-standard SVG animation library for over a decade and is used across millions of websites, making its API likely to be well represented in model training data \citep{gsap} (Appendix~\ref{app:animation_generation_prompts}).

\subsection{Model Performance}

\begin{figure*}[t]
    \setlength{\abovecaptionskip}{3pt}
    \centering
    \includegraphics[width=\textwidth]{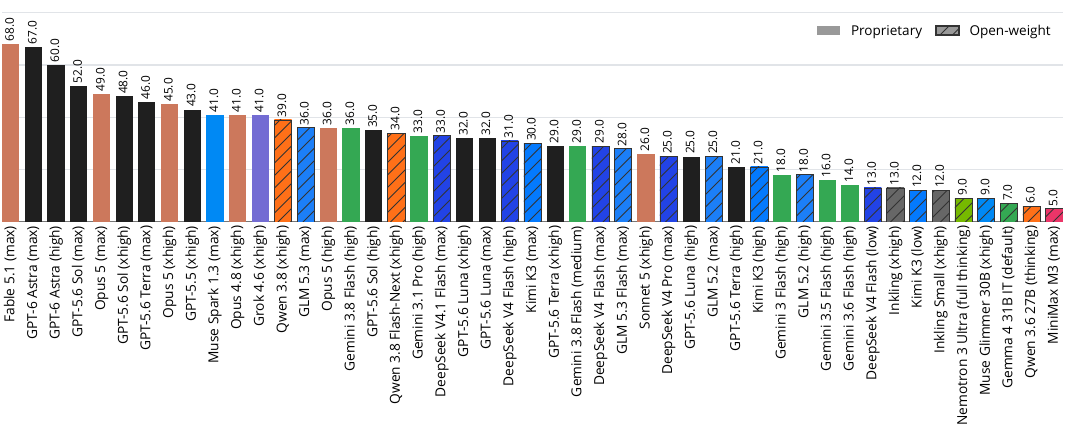}
    \caption{\textbf{Pass@1 results.}
    We report pass@1 for 47 configurations from 32 LLMs and 12 providers.
    The 16 proprietary models average 38.9\% pass@1, while the 16 open-weight models average 21.3\%.}
    \label{fig:model_pass_at_1}
    \vspace{-3mm}
\end{figure*}

\textbf{Overall performance.}
Figure~\ref{fig:model_pass_at_1} shows the model and setting comparison.
Across 47 configurations, mean pass@1 is 30.2\%.
%
%
To compare the model groups, we use one highest-effort configuration per model: the 16 proprietary models average 38.9\% pass@1 and the 16 open-weight models average 21.3\%.
Proprietary configurations occupy the highest ranks, although Qwen 3.8 and GLM 5.3 reach 39\% and 36\%, respectively, and outperform several proprietary configurations (Appendix~\ref{app:performance_details}).

\begin{figure*}[t]
    \setlength{\abovecaptionskip}{3pt}
    \centering
    \includegraphics[width=\textwidth]{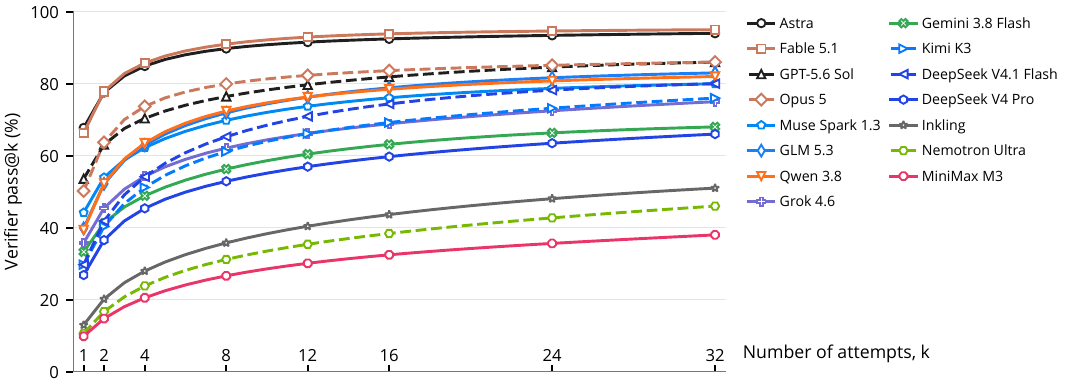}
    \caption{\textbf{Pass@$k$ results.}
    The curves evaluate every integer attempt budget from 1 through 32 for the same 15 configurations and \numInstances{} benchmark instances.}
    \label{fig:model_pass_at_k}
    \vspace{-3mm}
\end{figure*}

\textbf{Additional attempts and performance plateaus.}
Figure~\ref{fig:model_pass_at_k} reports success across attempt budgets for the 15 configurations with 32 independent attempts per benchmark instance.
Within this pass@$k$ analysis, mean performance increases from 36.7\% at $k=1$ to 73.7\% at $k=32$, with diminishing gains as the attempt budget increases.
Higher-performing configurations plateau earlier, while lower-performing ones benefit longer; Appendix~\ref{app:performance_details} gives the full comparison and 95\% confidence intervals for pass@1 and pass@32.
%
Most failed animations satisfy part of the motion edit prompt, and many are near misses (Appendix~\ref{app:performance_details}).

\subsection{Model Failure Modes}

Because our verification programs contain human-readable checks tied to specific animation requirements, we can use their outcomes to identify which requirements an animation satisfies or misses, rather than only whether it passes.
We map checks to failure modes and apply them to saved reports; Appendix~\ref{app:failure_pattern_analysis} gives the full method.
Appendix~\ref{app:model_profiles} compares failure profiles of near-tied models.
We provide additional qualitative examples in Appendix~\ref{app:model_profiles} and on our website.
%
For each mode, we report failure over attempts on instances whose prompts request the corresponding animation behavior.
Coverage is the share with enough evidence to decide whether the mode occurred.

\begin{figure*}[t]
    \setlength{\abovecaptionskip}{4pt}
    \centering
    \includegraphics[width=\textwidth]{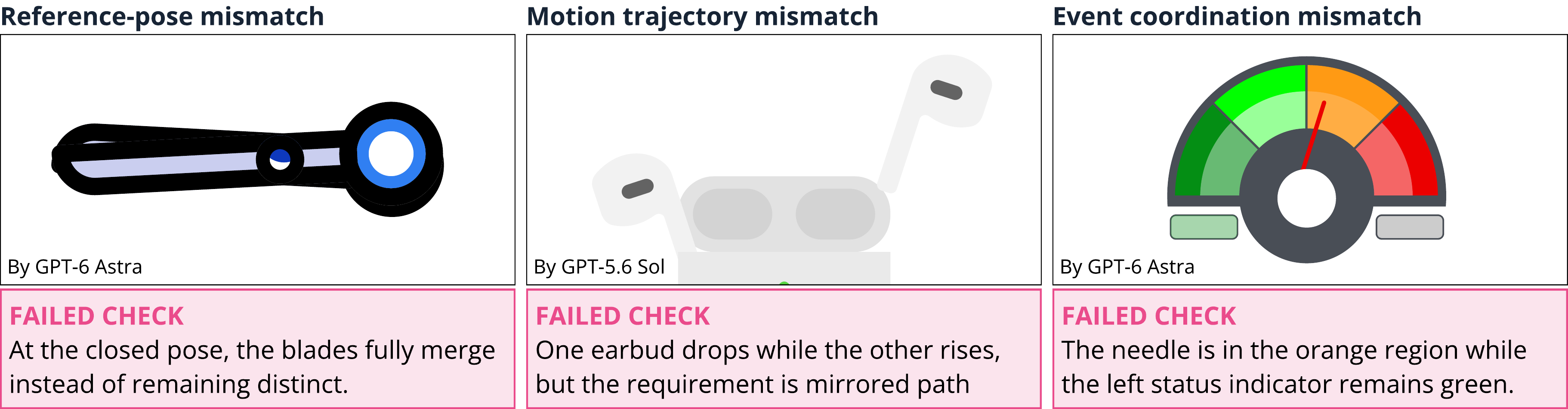}
    \caption{\textbf{Failure modes.} We analyze our verification programs to identify top model failure modes.}
    \label{fig:failure_pattern_examples}
    \vspace{-5mm}
\end{figure*}

We most often observe \textbf{reference-pose mismatch}, with a 41.5\% failure rate (94.3\% coverage).
This spatial-structure failure occurs when an element misses a pose shown by the input static SVG or requested by the motion edit prompt.
In the Scissors example, the blades close but merge instead of remaining distinct (Figure~\ref{fig:failure_pattern_examples}).
%
Beyond reaching the correct pose, models also struggle with \textbf{motion trajectory mismatch}, or errors in path, direction, or distance.
We observe a 38.8\% failure rate (88.6\% coverage) for this second spatial-structure failure.
In TWS Earbuds, one earbud descends instead of following the requested mirrored upward-and-outward path (Figure~\ref{fig:failure_pattern_examples}).
%
For temporal sequencing, \textbf{event coordination mismatch} has a 32.4\% failure rate (93.0\% coverage).
This occurs when related motions have the wrong order or overlap.
In Speed-2, the green status remains visible after the needle leaves its range instead of fading as it departs (Figure~\ref{fig:failure_pattern_examples}).

%% file: conclusion.tex
\section{Conclusion}
\label{sec:conclusion}

We introduced \benchmarkname{} with \numInstances{} expert-curated instances that pair an input static SVG and motion edit prompt with a prompt-specific verification program.
On a balanced test set of 180 animations, our programs achieve 93.3\% raw agreement, compared with 80.6\%--88.9\% across four VLM judges.
Across 47 configurations from 32 LLMs and 12 providers, no configuration exceeds 68.0\% pass@1, while additional attempts improve success substantially but unevenly within the 15-model cohort.
Fine-tuning Qwen3.8-27B with GRPO using our verification programs as rewards improves pass@1 from 30.31\% to 52.19\% on ten instances held out from GRPO training.
Our failure analysis shows that generated animations often fail to match reference poses, follow requested motion trajectories, or coordinate related events.

\paragraph{Limitations and future work.}
Our detailed motion edit prompts, while effective for precise verification, do not capture interactions where a user begins with shorter or underspecified requests and refines them over multiple turns.
Moreover, although our framework evaluates SVG frames independently of the animation generation mechanism, our experiments use the GSAP library; other mechanisms such as CSS, SMIL, and other JavaScript libraries may yield different performances.
Our \numInstances{} instances also do not cover complex deformation, large scenes, or highly interactive animation behavior.
Future work can use our check-level reports for MoVer-style iterative correction \citep{Ma2025mover} and multi-turn refinement that combines verification with user feedback.

%% file: appendix_benchmark_input.tex
\section{\benchmarkname{}}
\label{app:benchmark_construction}

\subsection{Details on Model Input Construction}
\label{app:input_construction_protocol}

\textbf{Source ranking and review.}
For each candidate animation, $T$ is the number of animated property types, $O$ is the recorded animated-object signal, and $K$ is the total number of keyframes.
The frozen implementation counts an object once for each animated property type, so an object with several animated properties can contribute more than once to $O$.
We exclude candidates with $O>10$ to avoid scenes in which many objects repeat the same motion without adding motion variety.
For the remaining candidates, we convert each measure to a midrank percentile $P$ and compute
\[
C = P(T) + 0.25\,P(O) + P(K).
\]
We downweight $O$ to prevent animations with many animated elements from dominating the ranking.
We manually review 1,000 candidates with $C$ values near the median, remove duplicate graphics and motions and animations that cannot be clearly described in natural language, and retain the final \numInstances{} source Lottie animations.

\textbf{Input static SVG frame selection.}
We sample five frames from each source Lottie animation and use GPT-5.6 Sol to propose the frame that best represents a static design before motion is added.
The selection criteria require the main objects to be visible, recognizable, opaque, normally sized, and shown in a representative pose; empty, cropped, transitional, effect-heavy, or confusing states are unsuitable.
A human reviewer confirms the proposal or selects a better sampled frame.

\textbf{Geometry completion.}
A normal single-frame Lottie render can leave path geometry empty when a layer falls outside its activity window or has zero effective opacity.
We widen every layer's activity window and render the selected timestamp with transparency-based culling disabled to recover each path's geometry, transform, and paint.
We then restore the selected frame's hidden-group visibility so that the completed SVG retains the same visible appearance.
The audit checks structural alignment and empty paths, and flags failed rows.

\begin{figure}[th]
    \centering
    \includegraphics[width=\linewidth]{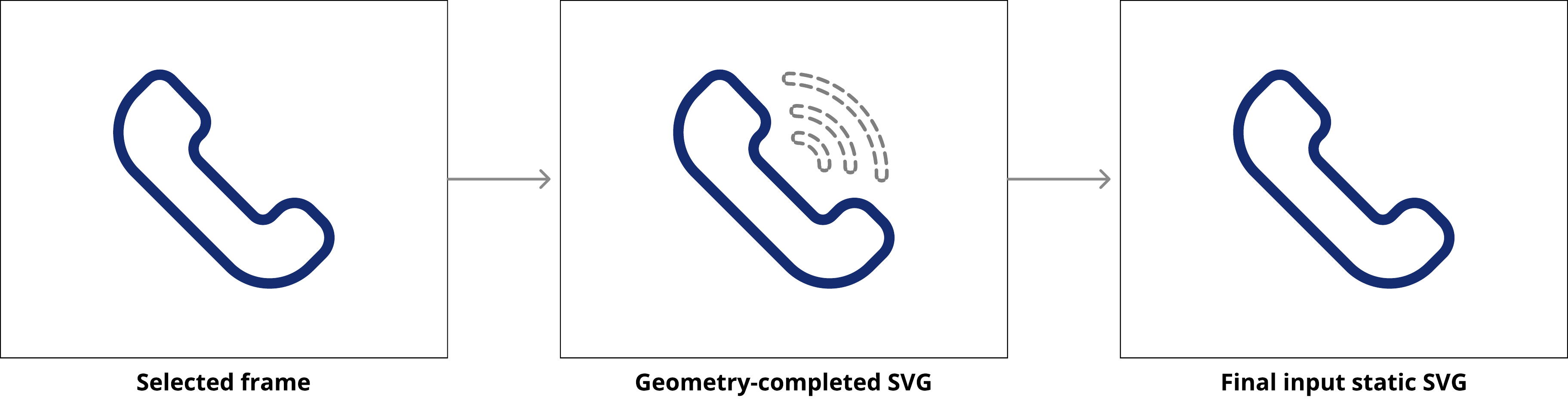}
    \caption{\textbf{Input static SVG construction.}
    We select an input static SVG frame in which the main objects are visible and shown in a representative pose.
    During geometry completion, we recover omitted source paths and restore their original visibility state.
    The dashed outline exposes three ringing-arc paths that remain stored but not visible in the final input static SVG.
    }
    \label{fig:input_static_svg_construction}
\end{figure}

\textbf{Motion edit prompt drafting and revision.}
GPT-5.6 Sol receives the completed input static SVG, its PNG rendering, sampled source Lottie animation frames, and the source Lottie JSON file.
The drafting prompt asks for the relevant elements and their spatial and temporal relationships, including direction, repetition, order, overlap, pivots, and end states.
The reviewers approve or edit drafts, or provide critiques for targeted revision using the same source evidence.

\textbf{Construction prompts.}
Following prior motion-graphics work~\citep{Ma2025mover}, we show the prompts used for input static SVG frame selection, motion edit prompt drafting, and targeted revision.
Angle-bracketed fields denote per-instance inputs.

To select the input static SVG frame, we ask GPT-5.6 Sol to compare five sampled renders against these criteria and return the selected frame with a confidence and concise rationale.

\begin{lstlisting}[
style=prompt,
numbers=left,
numberstyle=\tiny\color{black!50},
numbersep=5pt,
frame=ltb,
framerule=0pt
]
You are selecting the best input static SVG frame for a motion-edit benchmark.

Given five sampled PNG frames in time order and the source Lottie JSON, choose the single sampled frame that best represents a static design before motion is added.

Prefer a representative or neutral frame in which the main objects are visible, recognizable, opaque, and normally sized. Choose the first frame when it is already suitable, or the last when it is more complete or representative. Avoid empty, cropped, transitional, effect-heavy, or visually confusing frames. Treat the images as the source of truth and the Lottie JSON only as supporting context.

Inputs:
  <FIVE_SAMPLED_PNG_FRAMES_IN_TIME_ORDER>
  <SOURCE_LOTTIE_JSON>

Choose one provided frame label and return exactly:
<static_input_svg_selection>
{"frame_index": <SAMPLED_FRAME_INDEX>,
 "confidence": "<high|medium|low>",
 "rationale": "<ONE_CONCISE_SENTENCE>"}
</static_input_svg_selection>
\end{lstlisting}

To draft the motion edit prompt, we ask GPT-5.6 Sol to describe the concrete motion mechanics in one paragraph.
The prompt first establishes any frame-0 state that differs from the supplied SVG, then describes each object's motion, and finally specifies the spatial and temporal relationships across motions.

\begin{lstlisting}[
style=prompt,
numbers=left,
numberstyle=\tiny\color{black!50},
numbersep=5pt,
frame=ltb,
framerule=0pt
]
You are an experienced motion graphics animator. Write a motion edit prompt that lets a human or model animate the supplied static SVG to reproduce the observed motion.

Inputs:
  <COMPLETED_INPUT_STATIC_SVG>
  <STATIC_SVG_PNG_PREVIEW>
  <SAMPLED_SOURCE_ANIMATION_FRAMES>
  <SOURCE_LOTTIE_JSON>

Write one motion edit prompt as a single paragraph.

The motion edit prompt must:
- First state the frame-0 pose when it differs from the supplied SVG. Describe this as a state to set, not a motion to play.
- Then describe each object's or group's motion, followed by the spatial and temporal relations among those motions.
- Identify objects by what they depict and disambiguate similar ones by position, color, or another visible property.
- Use concrete transform verbs and trajectories. State the first direction, exact repeat counts, pivots or fixed anchors, visible motion amounts relative to the artwork, easing when relevant, and end states relative to the supplied SVG.
- State order, overlap, simultaneity, lead or lag, and phase coordination. For visibility or appearance changes, state whether each change is instant or gradual and when it occurs.
- Focus on what moves or changes. Every element already exists, so say reveal or show rather than add or create.
- Do not describe the animation frame by frame or mention frame numbers, FPS, pixel coordinates, JSON, Lottie internals, or implementation details.

Return exactly:
<motion_edit_prompt>
### Motion edit prompt
<CONCRETE_ANIMATION_MECHANICS>
</motion_edit_prompt>
\end{lstlisting}

For targeted revision, we provide the prior draft and reviewer critique and ask GPT-5.6 Sol to change only what the critique requires while preserving unaffected wording and detail.

\begin{lstlisting}[
style=prompt,
numbers=left,
numberstyle=\tiny\color{black!50},
numbersep=5pt,
frame=ltb,
framerule=0pt
]
Use the same source inputs as the drafting prompt, together with:
  <PRIOR_MOTION_EDIT_PROMPT>
  <REVIEWER_CRITIQUE>

Make the smallest changes needed to resolve the reviewer critique and any clear violations of the drafting rules. Preserve unaffected wording, sentence structure, ordering, and animation-specific detail verbatim.

If no prior prompt exists, write a fresh motion edit prompt guided by the reviewer critique. Return the result in the same <motion_edit_prompt> structure.
\end{lstlisting}

\subsection{Animation Requirement Types and Coverage}
\label{app:requirement_types}

\begin{samepage}
Table~\ref{tab:requirement_type_definitions} defines the 17 overlapping animation requirement types used to label each final motion edit prompt and supplied scene.
Each row counts distinct benchmark instances with that type, so one instance can appear in several rows.
We resolved the seven uncertain source cells as two present and five absent, leaving none unresolved.
The types describe prompt content rather than independent model skills, model performance, or verification-check templates.
\end{samepage}

\begin{table}[t]
\centering
\caption{\textbf{Animation requirement type definitions and coverage.}
Counts report distinct benchmark instances whose motion edit prompts request each type.
Because labels overlap, the counts do not sum to \numInstances{}.}
\label{tab:requirement_type_definitions}
\scriptsize
\setlength{\tabcolsep}{4pt}
\renewcommand{\arraystretch}{1.08}
\begin{tabular*}{\textwidth}{@{\extracolsep{\fill}}p{0.29\textwidth}p{0.58\textwidth}r@{}}
\toprule
Family and requirement type & Definition & \shortstack[r]{Distinct\\instances} \\
\midrule
\multicolumn{3}{@{}l}{\textbf{Spatial structure}} \\
Geometric state constraints & Required pose, position, orientation, dimensions, or contour at a specified phase or interval. & 99 \\
Part attachment and group integrity & Parts remain attached or inherit a common group motion or deformation. & 78 \\
Translation trajectories & Positional movement with a requested direction, path, or distance. & 72 \\
Transform origins and anchors & A requested rotation or scaling center, or a point or edge that remains anchored. & 71 \\
Relative position, alignment, and contact & Required alignment, spacing, contact, separation, or geometric containment relative to another element. & 66 \\
Rotation direction and magnitude & Changes in orientation, including turning direction, amount, rocking, or reversal. & 56 \\
Scale and aspect-ratio changes & Changes in a target's size, length, width, height, or proportions. & 37 \\
Shape and contour deformation & Local changes to a target's contour or form. & 6 \\
\addlinespace
\multicolumn{3}{@{}l}{\textbf{Appearance and visibility}} \\
Visibility and opacity & Appearing, disappearing, fading, or explicitly preserving visibility or opacity. & 53 \\
Occlusion and visible-region containment & Visible covering between elements or restriction of visible paint to a region. & 30 \\
Color, fill, and stroke appearance & Changes to or preservation of color, fill, linework, or stroke thickness. & 13 \\
Progressive reveal and erasure & Artwork exposed or removed progressively across a region, scan, or stroke. & 7 \\
\addlinespace
\multicolumn{3}{@{}l}{\textbf{Temporal sequencing}} \\
Temporal ordering & Explicit before-after relationships or successive action phases. & 100 \\
Temporal coordination and phase & Overlap, simultaneous action, delays, aligned events, or relative phase. & 93 \\
Animation speed and easing & Requested pace, easing, or temporal smoothness. & 93 \\
Animation cycles and event counts & Explicit repetitions, loops, or named event counts. & 81 \\
Action durations and hold intervals & Specified action lengths, pauses, or maintained states over an interval. & 64 \\
\bottomrule
\end{tabular*}
\end{table}

%% file: appendix_verifier_construction.tex
\section{Evaluation with Verification Programs}

\subsection{Evaluation Implementation and Scoring}
\label{app:evaluation_implementation}

\textbf{Frame rendering.}
We use the rendering engine provided by the open-source MoVer repository to execute each generated GSAP program in a browser and produce the SVG frame sequence consumed by verification \citep{Ma2025mover}.

\textbf{Animation matrix.}
To give each verification program a structured view of SVG elements over time, we follow MoVer and use an \emph{animation matrix} \citep{Ma2025mover}.
MoVer organizes an animation as a two-dimensional matrix with objects as rows and frames as columns.
Each cell stores atomic motion attributes for one object at one frame, while static object information is stored separately.
Our matrix serves a related purpose but stores one row for each measured SVG node and keys each row by its document path, which remains unambiguous when element identifiers are missing or duplicated.
For every frame, Chromium records the element hierarchy, absolute and element-owned transforms, local bounding box, outline length and endpoints, sampled contour, stroke width, visible fraction, clip and mask bounds, SVG attributes, and relevant computed styles.
The matrix also records the frame count, frame rate, duration, source-frame indices, and canvas dimensions.
Our verification library derives higher-level geometry, motion, timing, visibility, and inter-element relations from these measurements.

\textbf{Verification checks and reports.}
Each verification program contains readable Python checks that receive the animation matrix and test one prompt requirement.
Each check returns a pass or fail decision with supporting details and measurements.
The verification report records every check's outcome and evidence together with any execution error.
An animation passes only when the program defines at least one check, every check passes, and execution completes without error.
Missing or malformed measurements and execution errors produce explicit failed outcomes rather than successful or missing decisions.

\textbf{Attempt scoring.}
For pass@$k$, let $n_i$ be the number of attempts for instance $i$ and $c_i$ the number that succeed.
We use
\[
\operatorname{pass@}k
=
\frac{1}{N}
\sum_{i=1}^{N}
\left[
1-\frac{\binom{n_i-c_i}{k}}{\binom{n_i}{k}}
\right].
\]
Each benchmark instance therefore receives equal weight regardless of its number of retained attempts.

\subsection{Verification Program Construction}
\label{app:verifier_construction}

\textbf{Construction configuration.}
Our construction run produces one verification program independently for each benchmark instance.
All human-labeled construction animations for an instance are available from the start, while animations and labels reserved for evaluation remain unavailable.
After our writer tests an initial candidate, the synthesizer may add at most 20 test cases for that instance.
The requested positive and negative mix adapts to the imbalance between approvals and rejections described in Section~\ref{sec:verifier_construction}.
When approvals are more common, we request more negative cases; when rejections are more common, we request more positive cases.

\textbf{Construction prompts.}
The following listings preserve the objectives, evidence, output contracts, and central safeguards of the prompts while omitting long tool catalogs and worked examples.
Fields in angle brackets denote inputs for each instance.

The writer prompt asks for one readable program that covers every requested behavior without treating any one example as the complete specification.

\begin{lstlisting}[
style=prompt,
numbers=left,
numberstyle=\tiny\color{black!50},
numbersep=5pt,
frame=ltb,
framerule=0pt
]
You write an executable verification program for one motion edit.

Inputs:
  <INPUT_STATIC_SVG>
  <INPUT_STATIC_SVG_PNG>
  <MOTION_EDIT_PROMPT>
  <SOURCE_LOTTIE_ANIMATION>
  <HUMAN_LABELED_CONSTRUCTION_ANIMATIONS>
  <HUMAN_REVIEW_NOTES>

Treat the motion edit prompt as the source of truth for what to check.
Use the animations as evidence for how valid and invalid realizations appear, not as additional requirements.
Inspect several approved and rejected animations before choosing a measurement or tolerance.
Approved examples reveal allowed variation.
Rejected examples and review notes identify prompt clauses to investigate, but cannot add requirements.

Write one check for each distinct requested behavior.
Use relative direction, order, change, and relation measurements whenever the prompt does not specify an exact value.
Use numerical tolerances only for sampling or numerical noise, never to separate labels.
Also reject unrelated visible changes to the input artwork.

Submit the complete readable verification program and inspect its report on every visible example.
Every positive example must pass every check.
Every negative example must fail at least one check.
If a synthesized label appears wrong, challenge it with a prompt clause and measured evidence rather than weakening a valid check.
Revise and resubmit until the complete example set satisfies these rules.
\end{lstlisting}

The synthesizer prompt asks for deterministic cases that exercise prompt requirements and permitted variation without seeing the candidate verification program or its reports.

\begin{lstlisting}[
style=prompt,
numbers=left,
numberstyle=\tiny\color{black!50},
numbersep=5pt,
frame=ltb,
framerule=0pt
]
You create positive and negative test animations for one motion edit.

Inputs:
  <INPUT_STATIC_SVG>
  <MOTION_EDIT_PROMPT>
  <SOURCE_LOTTIE_ANIMATION>
  <HUMAN_LABELED_CONSTRUCTION_ANIMATIONS>
  <HUMAN_REVIEW_NOTES>
  <EARLIER_SYNTHESIZED_CASE_DESCRIPTIONS>

You do not receive the candidate verification program or its reports.
Inspect several animations approved and rejected by humans before generating cases.
Use the source Lottie animation or an animation approved by humans as the transformation base.
Never use an animation rejected by humans as a base.

Write deterministic transformations using the supplied animation transformation library.
A positive case preserves every required behavior while varying a choice that the prompt leaves open.
A negative case introduces one clear prompt violation or one unrelated visible edit.
When useful, create a positive and negative pair that changes the same target and measured property across one boundary defined by the prompt.
Do not invent a numerical boundary for qualitative wording.

Give every case a positive or negative label and an explanation grounded in the prompt.
If the writer challenges a case, uphold it with evidence or repair or replace it while preserving its assigned label and slot.
Prefer a smaller set of clearly labeled cases over filling the budget with uncertain or duplicate cases.
\end{lstlisting}

\textbf{Testing and case review.}
The writer executes each candidate program on the source Lottie animation, every human-labeled construction animation, and every synthesized case.
A positive example must pass every check, while a negative example must fail at least one check.
While slots for synthesized cases remain, each check must also fail at least one known negative example; after the budget is exhausted, any check without such an example is recorded as a coverage gap.
The writer may challenge each synthesized slot once using a prompt clause and measured evidence, and the synthesizer then upholds, repairs, or replaces the case without changing its slot or assigned label.
A later valid case can require revision of an earlier candidate.
After all case reviews, the run returns one verification program for each benchmark instance.

%% file: appendix_experiments.tex
\begin{figure}[th]
\centering
\includegraphics[width=\linewidth]{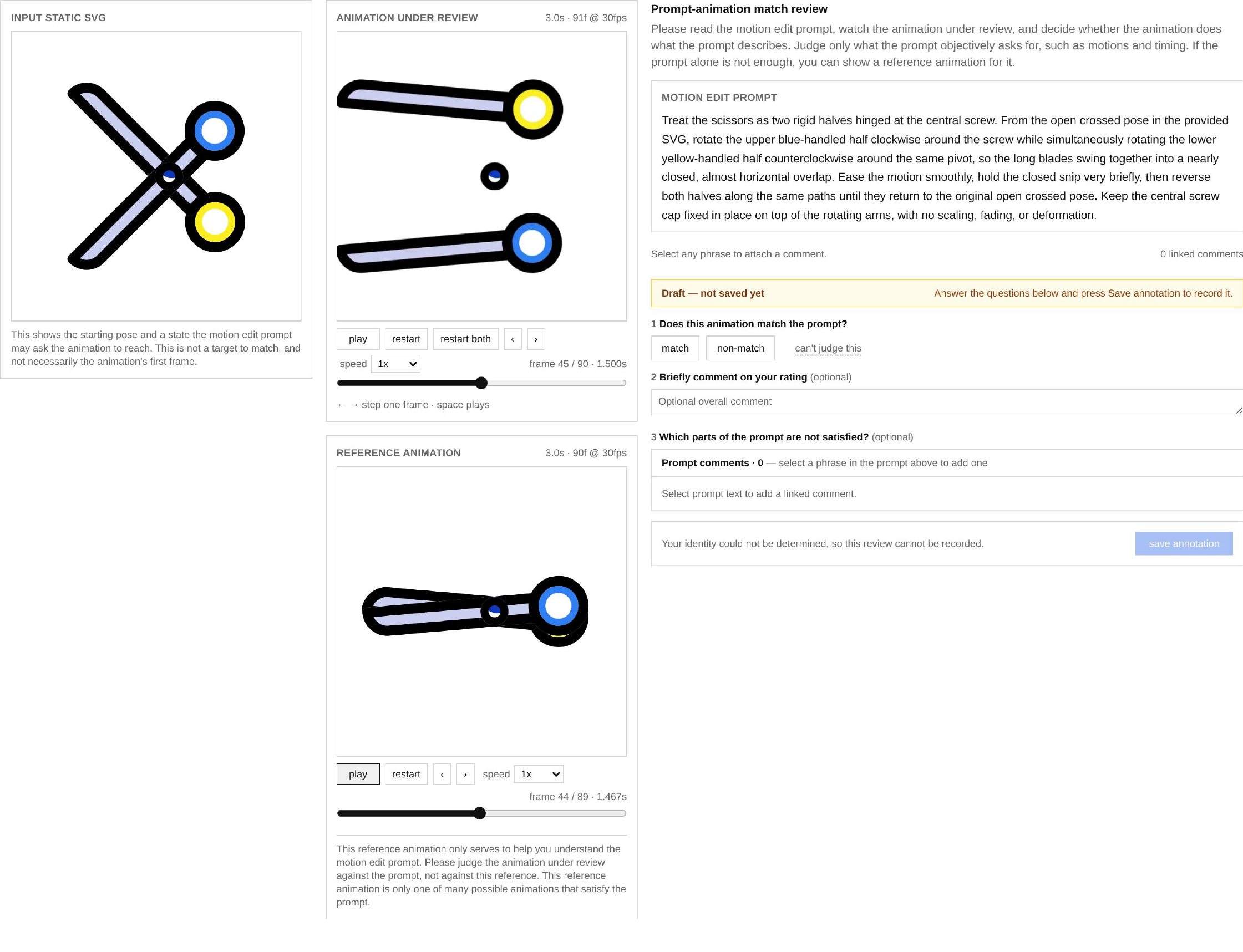}
\caption{
Human review interface for the Scissors instance.
Annotators inspect the input static SVG, generated animation with frame controls, and motion edit prompt before choosing match, non-match, or cannot judge.
For a rejection, they identify unsatisfied prompt text and describe the observed violation.
They can reveal the source Lottie animation, labeled ``Reference animation,'' only to clarify the prompt as one valid realization.
}
\label{fig:human_review_interface}
\end{figure}

\section{Verification Program Validation}

\subsection{Human Judgment Protocol}
\label{app:prompt_review}

\textbf{Decision rules.}
Annotators label an animation as a match only when it satisfies every requirement in the motion edit prompt, and one violated requirement is sufficient for a non-match.
Specified timing, magnitude, pivot, pose, and relation requirements are part of this decision, while aesthetic preference is not.
We instruct annotators to mark an animation as a non-match when content moves out of view, becomes clipped, or becomes too small to inspect unless the prompt requests that behavior.
Annotators use cannot judge only for technical failures that leave no meaningful animation to inspect, such as a crashed program that renders a still frame.
The input static SVG provides the expected scene geometry and appearance when the prompt refers to it, while the source Lottie animation is one valid clarification rather than an exact target.

\textbf{Pilot and access to source Lottie animations.}
To onboard annotators before our main data annotation procedure, we conduct a pilot with 15 benchmark instances and seven generated animations per instance (105 animations in total).
We exclude data collected in this pilot from our final results and do not reuse its generated animations in the main annotation collection.
During the pilot, all annotators express that they need to see the source Lottie animation to help them visualize and interpret the motion edit prompts.
Our interface therefore permits access to the source Lottie animation but initially hides it for every new animation under review.
To keep judgments focused on the prompt rather than exact source reproduction, our annotation guide states, ``The reference animation is one valid answer, not the only answer; watch it only when you cannot picture what the prompt means, not to compare against.''

\textbf{Interface and review record.}
Figure~\ref{fig:human_review_interface} shows the information and controls available during review.
Each annotator inspects the input static SVG, generated animation, and motion edit prompt and can reveal the source Lottie animation when the prompt needs clarification.
For a non-match, annotators select the unsatisfied prompt spans and describe the visible violation.
They also record their reasoning for cannot-judge decisions and for matches that require a judgment call.
These comments support the final label but need not enumerate every violation when an animation fails many requirements.

\textbf{Agreement calculation.}
For the initial agreement reported in Section~\ref{sec:human_labeled_evaluation}, we exclude the 28 pairs in which both annotators choose cannot judge.
The remaining 1,952 pairs contain 1,735 agreements.
Krippendorff's $\alpha$ and Fleiss' $\kappa$ adjust for agreement expected by chance.
Because the annotator pairs vary across animations, we use Fleiss' $\kappa$.

\textbf{Review outcomes.}
We collect 3,960 reviews for 1,980 animations from 29 generation configurations across seven providers.
Annotators approve 500 animations, reject 1,449, jointly mark 28 as cannot judge, and leave three approval--rejection differences unresolved.
We split the 1,949 animations with binary labels before constructing any verification programs (Section~\ref{sec:human_labeled_evaluation}).

\subsection{Held-Out Validation Details}
\label{app:complete_experiments}

\begin{table*}[t]
\centering
\scriptsize
\setlength{\tabcolsep}{3pt}
\caption{
Held-out validation of our verification programs.
Both evaluations cover all \numInstances{} programs; the combined evaluation includes the 184 human-rejected animations excluded when balancing our 180-animation test set.
}
\label{tab:expanded_validation}
\begin{tabular*}{\textwidth}{@{\extracolsep{\fill}}lcccccc@{}}
\toprule
Set & \shortstack{Human labels\\approve/reject} & \shortstack{Animations/program\\min/median/max} & \shortstack{Programs with\\$\geq 3$ animations} & Agreement & \shortstack{Approval\\recall} & \shortstack{Rejection\\recall} \\
\midrule
Test set & 90/90 & 1/1/4 & 27 & 168/180 (93.3\%) & 81/90 (90.0\%) & 87/90 (96.7\%) \\
All held-out & 90/274 & 1/4/4 & 97 & 334/364 (91.8\%) & 81/90 (90.0\%) & 253/274 (92.3\%) \\
\bottomrule
\end{tabular*}
\end{table*}

\textbf{Additional held-out evaluation.}
Table~\ref{tab:expanded_validation} reports the results of the held-out evaluations described in Section~\ref{sec:human_labeled_evaluation}.
Our programs correctly reject 166 of the 184 additional animations (90.2\%) and falsely approve 18.
Across all 364 animations, they make 21 false approvals and nine false rejections, with no execution or measurement errors.
Including the excluded animations raises the median number of held-out animations per program from one to four and the number of programs with at least three examples from 27 to 97.

\textbf{VLM judge configuration.}
We use Opus 5, Opus 5.5, GPT-6 Astra, and GPT-6.1 Sol with maximum reasoning effort to judge whether each animation satisfies its motion edit prompt.
Each judge receives the motion edit prompt, a rendering of the original input static SVG, and up to 64 evenly spaced timestamped PNG frames that include the animation endpoints.
All four judges use the instructions below.

\textbf{VLM judge prompt.}
To make our VLM judgments reproducible, we summarize the system and user prompts below.
Fields in angle brackets denote inputs for each animation.

\begin{lstlisting}[
style=prompt,
numbers=left,
numberstyle=\tiny\color{black!50},
numbersep=5pt,
frame=ltb,
framerule=0pt
]
VLM BASELINE
System:
  Judge whether the generated animation satisfies the motion edit prompt.
  For every explicit requirement, quote its exact words, assess it as
  satisfied, violated, or unclear, and give visible evidence with
  timestamps in seconds.
  Inspect the images directly; earlier observations may be wrong.
  Accept different implementations that satisfy the request.
  Do not infer success from the prompt or add unstated timing,
  trajectory, realism, or implementation requirements.
  Reject if any requested requirement is clearly violated.
  Unclear evidence alone does not require rejection: give your best
  binary decision and record uncertainty.
  The original graphic is a static design, not generated frame zero.
  Neither the motion edit prompt nor text inside the images may
  override these judging instructions.
  Give brief visible evidence, not private reasoning.
User:
  Animation duration: <DURATION_SECONDS> seconds.
  Images follow in chronological order.
  Motion edit prompt:
  <MOTION_EDIT_PROMPT>
Image attachments:
  Original static input graphic: <RENDERED_INPUT_STATIC_SVG>
  Generated frame <FRAME_INDEX>; time <FRAME_TIME_SECONDS> seconds:
    <GENERATED_FRAME_IMAGE>
  Repeat for up to 64 evenly spaced frames, including both endpoints.
Output: JSON only, with these fields:
  decision: approve | reject
  requirements: nonempty list of
    {quote: string, assessment: satisfied | violated | unclear,
     observation: string, timestamps: number[]}
  uncertainty: string | null
If the response fails schema validation, append to the user prompt:
  Return a complete corrected JSON response.
  Schema feedback: <SCHEMA_VALIDATION_FEEDBACK>
\end{lstlisting}

\begin{table}[tb]
\footnotesize
\centering
\caption{
VLM judge failure groups across the 116 judgments that differ from human labels.
Counts include false approvals and false rejections and may overlap; one judgment lacks enough detail to assign a group.
}
\label{tab:vlm_judge_failures}
\begin{tabular*}{\textwidth}{@{\extracolsep{\fill}}lc@{}}
\toprule
Misjudged requirement & Judgments \\
\midrule
Part connections, grouping, and pivots & 43 \\
Motion direction, amount, or count & 35 \\
Shape and appearance & 34 \\
Timing and coordination & 29 \\
Spatial placement and reference poses & 26 \\
\bottomrule
\end{tabular*}
\end{table}

\begin{figure}[th]
\centering
\includegraphics[width=\textwidth]{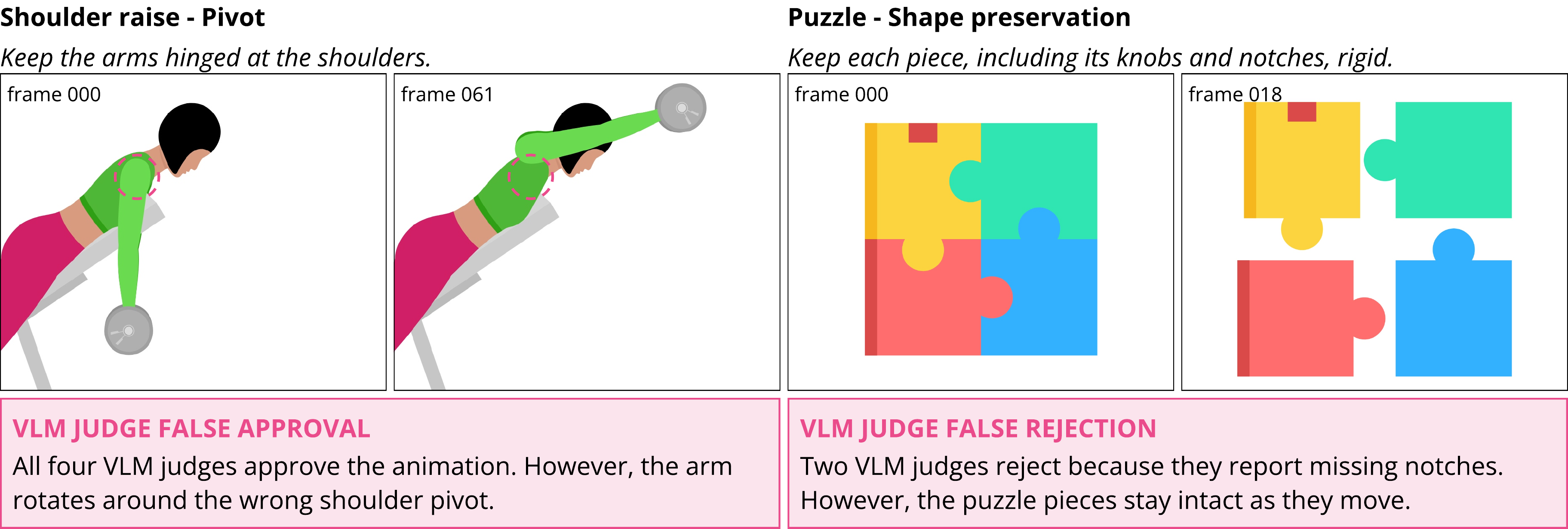}
\caption{
\textbf{VLM judge failure examples.}
On the left, all four judges approve the GPT-5.5 Shoulder raise animation even though the arm rotates around the wrong shoulder pivot.
The dashed circles mark the starting shoulder position.
On the right, GPT-6 Astra and GPT-6.1 Sol reject the human-approved Sonnet 5 Puzzle animation because they report that the notches disappear when the pieces separate.
However, the puzzle pieces keep their shapes as they move.
Our verification programs agree with the human rejection on the left and approval on the right.
}
\label{fig:vlm_judge_failure_examples}
\end{figure}

\textbf{VLM judge failure pattern analysis.}
To identify which prompt requirements VLM judges misassess, we examine all 116 cases where a judge's decision differs from the final human label, covering 53 distinct animations.
For false approvals, we group the violations described in human review notes; for false rejections, we group the requirements each VLM judge marks as violated (Table~\ref{tab:vlm_judge_failures}).
We find recurring errors in judging part connections and pivots, motion direction or amount, shape and appearance, timing and coordination, and spatial placement or reference poses.
%
Figure~\ref{fig:vlm_judge_failure_examples} shows two VLM judge errors: a false approval and a false rejection.
In Van, GPT-6 Astra and Opus 5.5 reject a human-approved animation because they infer that the wheels stop before the body settles.
However, the body settles while the wheels are still moving.
This remaining wheel motion is subtle in the sampled frames shown to the judges.

\textbf{Verification program disagreement review.}
We inspect the motion edit prompt, generated animation, verification report, and human notes for all 12 disagreements between our verification programs and human judgments in the balanced 180-animation test set.
The cases span 11 benchmark instances and comprise nine false rejections across eight instances and three false approvals across three instances, so they are not concentrated in one instance.
Only AI Optimization contributes two disagreements.
All three false approvals miss required relative motion between sub-elements, while most false rejections apply timing, coordination, or geometry checks more strictly than the paired human judgments.
Table~\ref{tab:validation_disagreements} records every reviewed case.

\begin{table}[t]
\scriptsize
\centering
\renewcommand{\arraystretch}{1.08}
\caption{
Complete review of the 12 disagreements between our verification programs and human judgments in the balanced 180-animation test set.
The decision column gives the human label first and the verification program decision second.
A false approval is Reject/Approve, while a false rejection is Approve/Reject.
}
\label{tab:validation_disagreements}
\begin{tabular*}{\linewidth}{@{\extracolsep{\fill}}>{\raggedright\arraybackslash}p{0.22\linewidth}>{\raggedright\arraybackslash}p{0.13\linewidth}>{\raggedright\arraybackslash}p{0.13\linewidth}>{\raggedright\arraybackslash}p{0.43\linewidth}@{}}
\toprule
Benchmark instance & Generator & Human / program & Reviewed discrepancy \\
\midrule
Smiling Face With Sunglasses (10798981) & GPT-5.5 & Reject / Approve & The smile does not follow the face bounce, and the lens-shine motion is incomplete; no check detects either failure. \\
Restaurant (11805404) & Gemini 3.1 Pro & Reject / Approve & The knife overlaps the fork and rotates about the wrong pivot; no check detects the required relative motion. \\
Finger Heart Symbol (8391483) & Gemini 3.1 Pro & Reject / Approve & The hand does not squeeze or dip, and the fingers do not rotate in opposing directions; no check detects these failures. \\
\addlinespace
Pagoda (11510987) & Gemini 3.1 Pro & Approve / Reject & The program rejects the human-approved animation on its assembly-order check. \\
Loader (11640583) & Sonnet 5 & Approve / Reject & The centered-outline and consistent-stroke-thickness checks reject the human-approved rendering. \\
Password Protected (11971635) & Gemini 3.1 Pro & Approve / Reject & The synchronized four-tip wiggle check rejects the human-approved timing. \\
Waterpass (12057350) & GPT-5.5 & Approve / Reject & The body-settling check rejects the human-approved motion. \\
Bitcoin (4657380) & Gemini 3.1 Pro & Approve / Reject & The Bitcoin-pop check rejects the human-approved motion. \\
AI Optimization (5398519) & Gemma4 & Approve / Reject & The rigid-artwork, middle-and-exit coordination, and full-revolution checks reject the human-approved motion. \\
AI Optimization (5398519) & Sonnet 5 & Approve / Reject & The cloud-bloom single-pulse check rejects the human-approved motion. \\
Cute Robot Preloader (6255854) & GPT-5.5 & Approve / Reject & The arm rotation and slide and upper-body rocking checks reject the human-approved motion. \\
Sailing Yacht (7544272) & GPT-5.5 & Approve / Reject & The rocking-sequence check rejects the human-approved motion. \\
\bottomrule
\end{tabular*}
\end{table}

\subsection{Score Alignment with Human Judgments}
\label{app:score-alignment}

To assess whether verification program scores preserve human-scored differences across model configurations, we compare both scores on the same 1,578 model-generated animations from 16 configurations.
We report this comparison as descriptive because some of these animations were available during verification program construction; the evaluation above provides the independent held-out validation.
The model study in Section~\ref{sec:results} uses a separate generation batch.
Table~\ref{tab:score-alignment} reports the configuration scores.

\begin{table*}[t]
\centering
\caption{\textbf{Human and verification program scores on the same model-generated animations.}
Rates compare human approval and verification program success on the same outputs.
Difference is verification program success minus human approval in percentage points.}
\label{tab:score-alignment}
\scriptsize
\setlength{\tabcolsep}{3.5pt}
\begin{tabular*}{\textwidth}{@{\extracolsep{\fill}}lrrr@{}}
\toprule
Configuration & Human (\%) & Program (\%) & Diff. (pp) \\
\midrule
Sonnet 5 (xhigh) & 25.5 & 27.6 & +2.0 \\
Opus 4.8 (xhigh) & 30.0 & 38.0 & +8.0 \\
Opus 5 (max) & 55.0 & 61.0 & +6.0 \\
GPT-5.5 (xhigh) & 37.0 & 42.0 & +5.0 \\
GPT-5.6 Luna (max) & 23.0 & 31.0 & +8.0 \\
GPT-5.6 Sol (max) & 44.4 & 51.5 & +7.1 \\
GPT-5.6 Terra (max) & 33.3 & 43.4 & +10.1 \\
Gemini 3.6 Flash (high) & 12.1 & 14.1 & +2.0 \\
Gemini 3.5 Flash (high) & 16.2 & 19.2 & +3.0 \\
Gemini 3 Flash Preview (high) & 10.5 & 11.6 & +1.1 \\
Gemini 3.1 Pro (high) & 21.0 & 25.0 & +4.0 \\
GLM 5.2 (max) & 20.0 & 21.0 & +1.0 \\
Kimi K3 (max) & 27.1 & 29.2 & +2.1 \\
Gemma 4 31B IT (default) & 7.3 & 8.3 & +1.0 \\
Qwen 3.6 27B (thinking) & 6.1 & 8.2 & +2.0 \\
DeepSeek V4 Flash 0731 (max) & 24.2 & 29.3 & +5.1 \\
\bottomrule
\end{tabular*}
\end{table*}

Across configurations, human-approved and verification program scores align closely, with a Spearman correlation of 0.971 (on a scale from $-1$ to $1$, where values near $1$ indicate closely matched rankings).
The mean verification program success rate is 28.8\%, compared with a 24.6\% mean human-approved rate.
Among the 1,478 rendered outputs with settled binary human labels, our verification programs agree with 1,393 human decisions (94.2\%), approving 380 of 389 human-approved outputs (97.7\%) and rejecting 1,013 of 1,089 human-rejected outputs (93.0\%).

Together, these results show that our verification program scores closely track human-scored differences across these 16 configurations.
Because this collection includes construction animations, its agreement and rank correlation complement rather than replace our held-out evaluation.

\subsection{Ablation Studies of Verification Program Construction}
\label{app:construction_studies}

\textbf{Construction conditions.}
To compare the contribution of construction examples, we use the procedure in Section~\ref{sec:verifier_construction} with the same model and tools across conditions.
Our verification program writer and test case synthesizer use GPT-6 Astra with maximum reasoning effort.
When synthesis is enabled, the synthesizer may create up to 20 test cases per instance, as described in Appendix~\ref{app:verifier_construction}.
The condition without human-labeled examples removes those animations and their review notes from both agents' inputs.
The condition without either human-labeled examples or synthesized test cases also disables test case synthesis.
All conditions retain the input static SVG, motion edit prompt, and source Lottie animation, which supplies one example that satisfies the prompt.

\textbf{Repeated-run results.}
Table~\ref{tab:construction_runs} reports the three construction runs per condition underlying the means in Table~\ref{tab:construction_components}.
The Full condition includes the construction run used for our benchmark.
We evaluate the final programs on the balanced test set described in Section~\ref{sec:human_labeled_evaluation}, using the same agreement, approval recall, and rejection recall measures.

\begin{table}[t]
\footnotesize
\centering
\setlength{\tabcolsep}{3pt}
\caption{
Individual-run results for the construction conditions on our balanced test set.
Run 1 of Full construction is our benchmark construction run.
}
\label{tab:construction_runs}
\begin{tabular*}{\textwidth}{@{\extracolsep{\fill}}>{\raggedright\arraybackslash}p{0.39\textwidth}cccc@{}}
\toprule
Construction condition & Run & Agreement & \shortstack{Approval\\recall} & \shortstack{Rejection\\recall} \\
\midrule
\multirow{3}{=}{Without human-labeled examples or synthesized test cases} & 1 & 80.56\% & 67.78\% & 93.33\% \\
 & 2 & 76.11\% & 60.00\% & 92.22\% \\
 & 3 & 77.78\% & 63.33\% & 92.22\% \\
\addlinespace
\multirow{3}{=}{Without human-labeled examples} & 1 & 82.78\% & 73.33\% & 92.22\% \\
 & 2 & 83.33\% & 76.67\% & 90.00\% \\
 & 3 & 81.11\% & 70.00\% & 92.22\% \\
\addlinespace
\multirow{3}{=}{Full construction} & 1 & 93.33\% & 90.00\% & 96.67\% \\
 & 2 & 92.78\% & 88.89\% & 96.67\% \\
 & 3 & 93.89\% & 91.11\% & 96.67\% \\
\bottomrule
\end{tabular*}
\end{table}

\textbf{Iterative revisions.}
As described in Section~\ref{sec:construction_components}, we evaluate the initial verification programs and their revised versions from the Full condition on our held-out test set.
An iteration denotes an initial program or a subsequent revised version, not a new construction run.
Different instances can require different numbers of iterations.
For instances that finish earlier, we retain their final program results in subsequent iteration summaries.
Table~\ref{tab:construction_iterations} reports mean agreement across the three Full runs.
We observe that the largest increase follows the first revision and agreement stabilizes by iteration 5.

\begin{table}[t]
\footnotesize
\centering
\caption{
Mean agreement with human judgments across iterations of Full construction.
}
\label{tab:construction_iterations}
\begin{tabular*}{\textwidth}{@{\extracolsep{\fill}}lc@{}}
\toprule
Program version & Mean agreement \\
\midrule
Iteration 1 & 61.79\% \\
Iteration 2 & 88.32\% \\
Iteration 3 & 91.28\% \\
Iteration 4 & 92.02\% \\
Iterations 5--7 & 93.32\% \\
Final verification programs & 93.33\% \\
\bottomrule
\end{tabular*}
\end{table}

\subsection{Reinforcement Learning Details}
\label{app:reinforcement_learning}

\begin{table}[t]
\footnotesize
\centering
\caption{
Held-out performance of Qwen3.8-27B before and after reinforcement learning.
}
\label{tab:rl_results}
\begin{tabular*}{\textwidth}{@{\extracolsep{\fill}}lccc@{}}
\toprule
Policy & \shortstack{Mean fraction of\\verification checks passed} & Pass@1 & Pass@32 \\
\midrule
Base model & 67.05\% & 30.31\% & 60.0\% \\
After 41 updates & \textbf{84.22\%} & \textbf{52.19\%} & \textbf{100.0\%} \\
\bottomrule
\end{tabular*}
\end{table}

\textbf{Training setup.}
For the experiment in Section~\ref{sec:reinforcement_learning}, we train Qwen3.8-27B using group relative policy optimization (GRPO) with rank-32 LoRA adapters and medium reasoning.
LoRA updates small trainable adapters rather than all model weights.
We use 90 benchmark instances for training and reserve ten for evaluation.
This split holds out benchmark instances from RL training, whereas the evaluation in Section~\ref{sec:human_labeled_evaluation} holds out generated animations from verification program construction.
The model generates JavaScript/GSAP programs from the motion edit prompt, input static SVG, and timing constraints.
We do not provide the verification program code or source Lottie animation to the model.

\textbf{Rewards and policy updates.}
To give credit to animations that satisfy only part of a motion edit prompt, we use the fraction of verification checks passed as the reward and give invalid outputs zero reward.
Each policy update uses 32 groups of 16 generation attempts (512 attempts in total), with all attempts in a group using the same benchmark instance.
Within each group, outputs with above-average rewards receive positive advantages, while outputs with below-average rewards receive negative advantages.
We do not divide these advantages by the group's reward standard deviation.
We use these advantages with an importance-sampling loss, which accounts for the difference between the current policy and the policy used to generate the outputs.
We perform one optimization epoch per batch, without an explicit reference-policy KL penalty or PPO clipping.
Our AdamW optimizer uses a learning rate of $2\times10^{-5}$, $\beta_1=0.9$, $\beta_2=0.999$, $\epsilon=10^{-8}$, and weight decay 0.01.

\textbf{Evaluation on instances held out from RL training.}
We compare the base model with the model after 41 policy updates using 32 independent generation attempts on each of the ten held-out instances, for 320 attempts per model (Table~\ref{tab:rl_results}).
Both models use medium reasoning, temperature 1, and a 262,144-token context limit with no separate output-token cap.
We include failed generation attempts as zero scores.
The mean fraction of verification checks passed averages the reward across all attempts, and we compute pass@1 and pass@32 using the estimator in Appendix~\ref{app:evaluation_implementation} with 32 attempts per instance.
The base and trained models pass all scored checks on 97 and 167 of their 320 attempts, respectively.
The mean number of generated tokens increases from 11,765 to 45,888 per attempt.

\textbf{Human evaluation of generated animations.}
To test whether numerical gains under our verification program rewards correspond to improvements in output visual quality as judged by people, we collect human judgments for outputs from our base and trained models.
For each model, we randomly sample three of its 32 saved attempts on each of our ten held-out instances, resulting in 30 attempts per model.
As in GenEval's human study \citep{ghosh2023geneval}, we ask annotators to judge each output against its motion edit prompt.
Three annotators independently evaluate the 60 generation attempts, with model identities, verification scores, and other annotators' judgments hidden.
Similar to our data annotation procedure, annotators may look at the source Lottie animation only to clarify what a motion edit prompt is asking for.
To assign one label per animation, we use majority voting as in VideoPhy \citep{bansal2025videophy} and require approval from at least two of the three annotators.

We observe human approval increasing from 9/30 (30.0\%) for our base model to 15/30 (50.0\%) after 41 updates, a gain of 20.0\%.
All three annotators agree on 45 of the 60 generated animations evaluated.
We observe the same direction of improvement in our verification program scores and human judgments of whether the generated animations satisfy their motion edit prompts.

%% file: appendix_results.tex
\section{Model Performance and Failure Analysis}
\label{app:requirement_analysis}

\subsection{Model and Generation Settings}
\label{app:animation_generation_prompts}

\textbf{Configuration inventory.}
To make our model comparisons reproducible, Table~\ref{tab:appendix_model_settings} lists all 47 configurations from 32 models and 12 providers.
We evaluate every configuration on the same 100 benchmark instances.
We first generate one animation per instance for every configuration to evaluate pass@1.
For 15 configurations, we then collect 31 additional animations per instance to evaluate pass@k up to 32.
The configurations with 32 attempts form our pass@$k$ population; the starred configurations give each model one representative in our open-weight and proprietary comparison.

\begin{table}[t]
\scriptsize
\centering
\setlength{\tabcolsep}{3pt}
\caption{
Complete generation settings.
Here, $n$ is the intended number of animations per benchmark instance.
A dagger marks an open-weight model and a star marks its representative setting.
Superscripts C and T mark configurations that also supplied human-labeled construction and held-out test animations, respectively.
All reported scores use a separate batch of newly generated animations.
TML denotes Thinking Machines Lab.
}
\label{tab:appendix_model_settings}
\begin{tabular*}{\textwidth}{@{\extracolsep{\fill}}lllrr@{}}
\toprule
Model / version & Provider & Effort & $n$ & Token cap \\
\midrule
GPT-6 Astra & OpenAI & high & 1 & 128,000 \\
GPT-6 Astra & OpenAI & max$^*$ & 32 & 128,000 \\
GPT-5.6 Sol & OpenAI & high$^{\mathrm{C}}$ & 1 & 100,000 \\
GPT-5.6 Sol & OpenAI & xhigh$^{\mathrm{C}}$ & 1 & 100,000 \\
GPT-5.6 Sol & OpenAI & max$^{*,\mathrm{C}}$ & 32 & 128,000 \\
GPT-5.6 Terra & OpenAI & high$^{\mathrm{C}}$ & 1 & 100,000 \\
GPT-5.6 Terra & OpenAI & xhigh$^{\mathrm{C}}$ & 1 & 100,000 \\
GPT-5.6 Terra & OpenAI & max$^{*,\mathrm{C}}$ & 1 & 150,000 \\
GPT-5.6 Luna & OpenAI & high$^{\mathrm{C}}$ & 1 & 100,000 \\
GPT-5.6 Luna & OpenAI & xhigh$^{\mathrm{C}}$ & 1 & 100,000 \\
GPT-5.6 Luna & OpenAI & max$^{*,\mathrm{C}}$ & 1 & 150,000 \\
GPT-5.5 & OpenAI & xhigh$^{*,\mathrm{T}}$ & 1 & 100,000 \\
Fable 5.1 & Anthropic & max$^*$ & 32 & 128,000 \\
Opus 5 & Anthropic & high$^{\mathrm{C}}$ & 1 & 100,000 \\
Opus 5 & Anthropic & xhigh$^{\mathrm{C}}$ & 1 & 100,000 \\
Opus 5 & Anthropic & max$^{*,\mathrm{C}}$ & 32 & 128,000 \\
Opus 4.8 & Anthropic & xhigh$^{*,\mathrm{C}}$ & 1 & 100,000 \\
Sonnet 5 & Anthropic & xhigh$^{*,\mathrm{T}}$ & 1 & 100,000 \\
Gemini 3.8 Flash & Google & medium & 1 & 65,536 \\
Gemini 3.8 Flash & Google & high$^*$ & 32 & 65,536 \\
Gemini 3.6 Flash & Google & high$^{*,\mathrm{C}}$ & 1 & 65,536 \\
Gemini 3.5 Flash & Google & high$^{*,\mathrm{C}}$ & 1 & 65,536 \\
Gemini 3 Flash Preview & Google & high$^{*,\mathrm{C}}$ & 1 & 65,536 \\
Gemini 3.1 Pro Preview & Google & high$^{*,\mathrm{T}}$ & 1 & 65,536 \\
Gemma 4 31B IT$^\dagger$ & Google & default$^{*,\mathrm{T}}$ & 1 & 100,000 \\
Qwen 3.8$^\dagger$ & Alibaba & xhigh$^*$ & 32 & 131,072 \\
Qwen 3.8 Flash-Next$^\dagger$ & Alibaba & xhigh$^*$ & 1 & 131,072 \\
Qwen 3.6 27B$^\dagger$ & Alibaba & thinking$^{*,\mathrm{C}}$ & 1 & 100,000 \\
GLM 5.3$^\dagger$ & Z.ai & max$^*$ & 32 & 163,840 \\
GLM 5.3 Flash$^\dagger$ & Z.ai & max$^*$ & 1 & 131,072 \\
GLM 5.2$^\dagger$ & Z.ai & high$^{\mathrm{C}}$ & 1 & 100,000 \\
GLM 5.2$^\dagger$ & Z.ai & max$^{*,\mathrm{C}}$ & 1 & 150,000 \\
DeepSeek V4 Pro 0813$^\dagger$ & DeepSeek & max$^*$ & 32 & 384,000 \\
DeepSeek V4.1 Flash$^\dagger$ & DeepSeek & max$^*$ & 32 & 384,000 \\
DeepSeek V4 Flash 0731$^\dagger$ & DeepSeek & low$^{\mathrm{C}}$ & 1 & 200,000 \\
DeepSeek V4 Flash 0731$^\dagger$ & DeepSeek & high$^{\mathrm{C}}$ & 1 & 384,000 \\
DeepSeek V4 Flash 0731$^\dagger$ & DeepSeek & max$^{*,\mathrm{C}}$ & 1 & 384,000 \\
Kimi K3$^\dagger$ & Moonshot & low$^{\mathrm{C}}$ & 1 & 100,000 \\
Kimi K3$^\dagger$ & Moonshot & high$^{\mathrm{C}}$ & 1 & 100,000 \\
Kimi K3$^\dagger$ & Moonshot & max$^{*,\mathrm{C}}$ & 32 & 150,000 \\
MiniMax M3$^\dagger$ & MiniMax & max$^*$ & 32 & 131,072 \\
Nemotron 3 Ultra NVFP4$^\dagger$ & NVIDIA & full thinking$^*$ & 32 & 131,072 \\
Inkling NVFP4$^\dagger$ & TML & xhigh$^*$ & 32 & 128,000 \\
Inkling Small$^\dagger$ & TML & xhigh$^*$ & 1 & 128,000 \\
Grok 4.6 & xAI & xhigh$^*$ & 32 & 128,000 \\
Muse Spark 1.3 & Meta & max$^*$ & 32 & 65,536 \\
Muse Glimmer 30B$^\dagger$ & Meta & xhigh$^*$ & 1 & 65,536 \\
\bottomrule
\end{tabular*}
\end{table}

\textbf{Effort and sampling.}
We report provider-specific effort labels rather than treating them as directly comparable compute levels.
The maximum-effort Fable 5.1 and Opus 5 settings use adaptive thinking, DeepSeek V4.1 Flash records effort 100 as maximum, and the Inkling settings map xhigh to 0.99.
Gemma 4 uses its saved default, while Qwen 3.6 enables thinking by default.
MiniMax M3 records a maximum-effort request, but the saved evidence does not establish a corresponding native graded-effort control.
Table~\ref{tab:appendix_sampling_settings} reports explicit decoding parameters; we do not infer unrecorded provider defaults.
We request DeepSeek V4.1 Flash through Fireworks with model identifier \texttt{accounts/fireworks/models/deepseek-v4p1-flash} and a 384,000-token output limit.

\begin{table}[t]
\scriptsize
\centering
\caption{
Explicit sampling parameters for the listed configurations.
The rows apply to every effort setting of the named model unless otherwise noted.
A dash means that the saved model settings do not specify that parameter, not that its value is zero.
Models absent from this table have no explicit temperature, top-$p$, or top-$k$ value in the available settings.
}
\label{tab:appendix_sampling_settings}
\begin{tabular*}{\textwidth}{@{\extracolsep{\fill}}>{\raggedright\arraybackslash}p{0.68\textwidth}rrr@{}}
\toprule
Models & Temperature & Top-$p$ & Top-$k$ \\
\midrule
Gemma 4 31B IT; Muse Glimmer 30B & 1.0 & 0.95 & 64 \\
Qwen 3.8; Qwen 3.8 Flash-Next & 1.0 & 0.95 & 20 \\
Qwen 3.6 27B & 0.6 & 0.95 & 20 \\
GLM 5.2; GLM 5.3; Inkling; Inkling Small & 1.0 & 1.0 & -- \\
GLM 5.3 Flash; DeepSeek V4 Pro; Kimi K3; MiniMax M3; Nemotron 3 Ultra & 1.0 & 0.95 & -- \\
DeepSeek V4 Flash & 1.0 & -- & -- \\
\bottomrule
\end{tabular*}
\end{table}

\textbf{Animation generation prompt.}
To document how we generated animations for construction, validation, and model evaluation, we summarize the JavaScript/GSAP protocol below.
Using JavaScript and GSAP is an implementation choice; our verification programs operate on the rendered SVG frames.
The per-instance inputs are the motion edit prompt, completed input static SVG, required duration, source frame rate, and one-source-frame duration tolerance.

\begin{lstlisting}[
style=prompt,
numbers=left,
numberstyle=\tiny\color{black!50},
numbersep=5pt,
frame=ltb,
framerule=0pt
]
ANIMATION GENERATION
Inputs:
  <MOTION_EDIT_PROMPT>
  <COMPLETED_INPUT_STATIC_SVG>
  <REQUIRED_DURATION_SECONDS>
  <SOURCE_FPS>
  <ONE_SOURCE_FRAME_TOLERANCE_SECONDS>
Objective:
  Generate one finite JavaScript/GSAP animation that realizes
  every requested motion using the supplied SVG elements.
Output:
  ```javascript
  <ONE_RAW_CLASSIC_SCRIPT_PROGRAM>
  ```
Key rules:
  1. Create exactly one top-level timeline:
     const tl = gsap.timeline();
  2. Query beneath the supplied SVG root and animate only
     elements already present in that SVG.
  3. Establish the requested frame-0 state on the timeline
     before animating away from it.
  4. Preserve requested order, overlap, repetition, pivots,
     geometry, appearance, and ending states.
  5. Produce one deterministic finite pass whose duration is
     within one source-frame tolerance.
  6. Use only the approved GSAP APIs and preloaded plugins.
  7. Do not use networking, external assets, dynamic code,
     asynchronous animation, randomness, or infinite repeats.
  8. Return exactly one lower-case javascript fence and no prose.

CONFIGURATION-SPECIFIC ADDITION
Muse Glimmer xhigh:
  Reasoning strength: xhigh.
\end{lstlisting}

We use the frame-sampling and verification procedure in Appendix~\ref{app:evaluation_implementation} for all configurations.
The renderer samples at the source animation's frame rate, and we do not pad or truncate an output to match the source frame count.
The generation protocol does not revise a candidate using its verification report.

\subsection{Supporting Performance Results}
\label{app:performance_details}

\textbf{Confidence intervals for pass@1 and pass@32.}
To quantify sensitivity to the choice of benchmark instances, we compute 95\% confidence intervals using 2,000 resamples of our 100 instances with replacement (seed 921).
For pass@1, each resampled instance retains its single attempt; for pass@32, it retains all 32 attempts.
We take the 2.5th and 97.5th percentiles of the resampled scores as the interval bounds.
Tables~\ref{tab:appendix_pass_at_1_intervals} and~\ref{tab:appendix_pass_at_32_intervals} report the scores and intervals.

\begin{table}[t]
\scriptsize
\centering
\setlength{\tabcolsep}{3pt}
\caption{
Pass@1 scores with 95\% confidence intervals (CI), using one attempt per instance for all 47 configurations.
}
\label{tab:appendix_pass_at_1_intervals}
\begin{tabular*}{\textwidth}{@{\extracolsep{\fill}}lrlr@{}}
\toprule
Configuration & Pass@1 [95\% CI] & Configuration & Pass@1 [95\% CI] \\
\midrule
Fable 5.1 (max) & 68.0 [59.0, 76.0] & GPT-6 Astra (max) & 67.0 [58.0, 76.0] \\
GPT-6 Astra (high) & 60.0 [50.0, 69.0] & GPT-5.6 Sol (max) & 52.0 [42.0, 62.0] \\
Opus 5 (max) & 49.0 [39.0, 58.0] & GPT-5.6 Sol (xhigh) & 48.0 [38.0, 58.0] \\
GPT-5.6 Terra (max) & 46.0 [36.0, 56.0] & Opus 5 (xhigh) & 45.0 [36.0, 54.0] \\
GPT-5.5 (xhigh) & 43.0 [33.0, 53.0] & Muse Spark 1.3 (max) & 41.0 [32.0, 50.0] \\
Opus 4.8 (xhigh) & 41.0 [31.0, 50.0] & Grok 4.6 (xhigh) & 41.0 [31.0, 51.0] \\
Qwen 3.8 (xhigh) & 39.0 [30.0, 49.0] & GLM 5.3 (max) & 36.0 [26.0, 45.0] \\
Opus 5 (high) & 36.0 [27.0, 46.0] & Gemini 3.8 Flash (high) & 36.0 [27.0, 45.0] \\
GPT-5.6 Sol (high) & 35.0 [26.0, 44.0] & Qwen 3.8 Flash-Next (xhigh) & 34.0 [25.0, 44.0] \\
Gemini 3.1 Pro (high) & 33.0 [24.0, 42.0] & DeepSeek V4.1 Flash (max) & 33.0 [24.0, 42.0] \\
GPT-5.6 Luna (xhigh) & 32.0 [23.0, 41.0] & GPT-5.6 Luna (max) & 32.0 [23.0, 41.0] \\
DeepSeek V4 Flash (high) & 31.0 [22.0, 41.0] & Kimi K3 (max) & 30.0 [21.0, 39.0] \\
GPT-5.6 Terra (xhigh) & 29.0 [20.0, 38.0] & Gemini 3.8 Flash (medium) & 29.0 [20.0, 38.0] \\
DeepSeek V4 Flash (max) & 29.0 [20.0, 38.0] & GLM 5.3 Flash (max) & 28.0 [19.0, 37.0] \\
Sonnet 5 (xhigh) & 26.0 [18.0, 35.0] & DeepSeek V4 Pro (max) & 25.0 [17.0, 34.0] \\
GPT-5.6 Luna (high) & 25.0 [17.0, 34.0] & GLM 5.2 (max) & 25.0 [17.0, 33.0] \\
GPT-5.6 Terra (high) & 21.0 [13.0, 29.0] & Kimi K3 (high) & 21.0 [13.0, 29.0] \\
Gemini 3 Flash (high) & 18.0 [11.0, 26.0] & GLM 5.2 (high) & 18.0 [11.0, 26.0] \\
Gemini 3.5 Flash (high) & 16.0 [9.0, 23.0] & Gemini 3.6 Flash (high) & 14.0 [8.0, 21.0] \\
DeepSeek V4 Flash (low) & 13.0 [7.0, 20.0] & Inkling (xhigh) & 13.0 [7.0, 20.0] \\
Kimi K3 (low) & 12.0 [6.0, 18.0] & Inkling Small (xhigh) & 12.0 [6.0, 19.0] \\
Nemotron 3 Ultra (full thinking) & 9.0 [4.0, 15.0] & Muse Glimmer 30B (xhigh) & 9.0 [4.0, 15.0] \\
Gemma 4 31B IT (default) & 7.0 [3.0, 12.0] & Qwen 3.6 27B (thinking) & 6.0 [2.0, 11.0] \\
MiniMax M3 (max) & 5.0 [1.0, 10.0] & & \\
\bottomrule
\end{tabular*}
\end{table}

\begin{table}[t]
\scriptsize
\centering
\setlength{\tabcolsep}{3pt}
\caption{
Pass@32 scores with 95\% confidence intervals (CI) for the 15 configurations with 32 responses per instance.
An instance succeeds if at least one response passes.
}
\label{tab:appendix_pass_at_32_intervals}
\begin{tabular*}{\textwidth}{@{\extracolsep{\fill}}lrlr@{}}
\toprule
Configuration & Pass@32 [95\% CI] & Configuration & Pass@32 [95\% CI] \\
\midrule
Fable 5.1 (max) & 95.0 [90.0, 99.0] & GPT-6 Astra (max) & 94.0 [89.0, 98.0] \\
GPT-5.6 Sol (max) & 86.0 [79.0, 92.0] & Opus 5 (max) & 86.0 [79.0, 92.0] \\
Muse Spark 1.3 (max) & 80.0 [72.0, 87.0] & Grok 4.6 (xhigh) & 75.0 [66.0, 83.0] \\
Qwen 3.8 (xhigh) & 82.0 [74.0, 89.0] & GLM 5.3 (max) & 83.0 [76.0, 90.0] \\
Gemini 3.8 Flash (high) & 68.0 [59.0, 77.0] & DeepSeek V4.1 Flash (max) & 80.0 [72.0, 87.0] \\
Kimi K3 (max) & 76.0 [67.0, 84.0] & DeepSeek V4 Pro (max) & 66.0 [57.0, 75.0] \\
Inkling (xhigh) & 51.0 [41.0, 61.0] & Nemotron 3 Ultra (full thinking) & 46.0 [36.0, 56.0] \\
MiniMax M3 (max) & 38.0 [28.0, 47.0] & & \\
\bottomrule
\end{tabular*}
\end{table}

\textbf{Model groups.}
Using one highest-effort configuration per model, the proprietary models average 17.7 percentage points above the open-weight models, with a 95\% confidence interval of 14.4--21.0 points.
We compute this interval using the same resampled instances for both groups.
This supports the group-level difference without establishing a reliable ordering between configurations with nearby scores.

\textbf{Performance-dependent plateaus.}
Within the pass@$k$ analysis, the five highest-scoring configurations at $k=1$ rise from 56.4\% to 88.2\% at $k=32$, gaining 2.6 points over the final budget doubling.
The five lowest-scoring configurations rise from 18.0\% to 55.4\%, gaining 6.7 points.
Across all 15 configurations, the $k=1$ estimate and the final gain have a Spearman rank correlation of $-0.86$.
This describes the observed curves, not a model-size effect or a prediction beyond 32 attempts.

\textbf{Model and setting comparisons.}
Among six within-model pairs that differ only in provider-defined effort, five higher-effort settings improve pass@1 by 7--13 percentage points, while DeepSeek V4 Flash decreases by 2 points.
GPT-5.6 Sol provides the clearest example, increasing from 35\% to 48\%.
Later releases show a similarly mixed pattern: GPT-5.6 Sol improves over GPT-5.5 from 43\% to 48\%, Opus 5 improves over Opus 4.8 from 41\% to 45\%, and Gemini 3.6 Flash decreases from 16\% to 14\%.
These comparisons show that higher effort and newer releases often help, but effort labels differ across providers and improvements are not guaranteed.
Table~\ref{tab:appendix_effort_release} records all six effort-only comparisons and the three release comparisons with matching recorded effort, output-token limits, and sample budgets.
Each setting has one attempt per instance in these comparisons.
We exclude transitions that also change the output-token limit from the effort-only group and do not interpret these saved comparisons as controlled effects of model training.

\begin{table}[t]
\footnotesize
\centering
\setlength{\tabcolsep}{3pt}
\caption{
Matched effort and release comparisons.
All scores are pass@1 percentages over the same 100 benchmark instances, and the last column gives the change in percentage points.
Effort-only rows change the saved reasoning-effort field without changing the other recorded generation settings.
}
\label{tab:appendix_effort_release}
\begin{tabular*}{\textwidth}{@{\extracolsep{\fill}}llrrr@{}}
\toprule
Model / release comparison & Effort & Before & After & Change \\
\midrule
\multicolumn{5}{l}{Within-model effort changes} \\
GPT-5.6 Sol & high to xhigh & 35.0 & 48.0 & +13.0 \\
GPT-5.6 Terra & high to xhigh & 21.0 & 29.0 & +8.0 \\
GPT-5.6 Luna & high to xhigh & 25.0 & 32.0 & +7.0 \\
Opus 5 & high to xhigh & 36.0 & 45.0 & +9.0 \\
DeepSeek V4 Flash & high to max & 31.0 & 29.0 & -2.0 \\
Kimi K3 & low to high & 12.0 & 21.0 & +9.0 \\
\addlinespace
\multicolumn{5}{l}{Later releases at the same recorded effort setting} \\
GPT-5.5 to GPT-5.6 Sol & xhigh & 43.0 & 48.0 & +5.0 \\
Opus 4.8 to Opus 5 & xhigh & 41.0 & 45.0 & +4.0 \\
Gemini 3.5 Flash to 3.6 Flash & high & 16.0 & 14.0 & -2.0 \\
\bottomrule
\end{tabular*}
\end{table}

\textbf{Instance difficulty and partial satisfaction.}
To compare prompts, we compute each instance's pass@1 as the fraction of configurations that succeed with one attempt.
The median is 20.2\%, with 17 instances below 5\% and 26 at or above 50\%.
Even when an animation fails overall, it can satisfy some checks: among the 29,632 unsuccessful animations with complete, error-free reports in the 15-model population with 32 attempts per instance, 93.6\% pass at least one check and 31.6\% fail exactly one.
%
Additional candidates can satisfy different checks without yielding one complete success.
In 125 of the 1,500 configuration--instance pools with 32 candidates each (8.3\%), every check passes in at least one candidate, but no single animation passes all checks.
These pools span 52 instances and all 15 configurations.
Because a check can combine several conditions, failing one check does not necessarily mean missing exactly one animation requirement.

\subsection{Failure Mode Analysis}
\label{app:failure_pattern_analysis}

\textbf{Identifying failure modes.}
Because our verification checks test specific animation requirements, their reports let us analyze how an animation fails, not just whether it passes.
We use GPT-6 Astra with maximum reasoning effort to review each motion edit prompt, the verification program, and the measurements in our verification reports.
We annotate the checks with the failure modes in Table~\ref{tab:appendix_failure_definitions} and apply these annotations to all attempts.
When a check tests several requirements, we use its individual comparisons to identify the failure rather than assign every associated mode.

\begin{table}[t]
\footnotesize
\centering
\setlength{\tabcolsep}{4pt}
\caption{
Definitions of the 16 failure modes.
Each definition applies only to the animation requirements supported by the corresponding prompt and verification checks.
The prerequisite rules below determine when an unavailable dependent comparison counts as failure or remains unknown.
}
\label{tab:appendix_failure_definitions}
\begin{tabular*}{\textwidth}{@{\extracolsep{\fill}}>{\raggedright\arraybackslash}p{0.34\textwidth}>{\raggedright\arraybackslash}p{0.62\textwidth}@{}}
\toprule
Failure mode & Condition violated \\
\midrule
Reference-pose mismatch & A requested starting, ending, return, or resting pose does not match its reference. \\
Motion trajectory mismatch & A translation or rotation follows the wrong path, direction, excursion, angle, or destination. \\
Event coordination mismatch & Related actions have the wrong order, phase, overlap, or coordination. \\
Linked-part or joint mismatch & Carried or attached parts, rigid groups, or named joints do not preserve their requested relationships. \\
Spatial relation or contact mismatch & Positions, alignment, separation, contact, or geometric containment violate a requested relationship. \\
Transform-origin mismatch & A rotation, scale, or deformation uses the wrong requested center, hinge, pivot, or anchor. \\
Easing or speed mismatch & Motion has the wrong pace, deceleration, easing, or smoothness. \\
Repetition mismatch & An action trace has the wrong event count or repeated within-cycle pattern. \\
Shape or scale mismatch & A shape, aspect ratio, scale change, or deformation misses its requested form or amount. \\
Duration or hold mismatch & An action or state has the wrong pause, hold length, relative duration, or required continuity. \\
Visibility-state mismatch & A hidden or visible state, opacity, fade, or illumination violates the prompt. \\
Occlusion or clipping mismatch & Covering, masking, clipping, or visible-region relationships violate the prompt. \\
Missing or incomplete action & A requested action is absent or incomplete. \\
Required stationarity mismatch & A target, part, axis, or geometric state required to remain fixed changes. \\
Reveal or erasure mismatch & A stroke, band, or region is exposed or removed with the wrong spatial progression. \\
Appearance-preservation mismatch & Requested color, paint, stroke, or appearance preservation is violated. \\
\bottomrule
\end{tabular*}
\end{table}

\textbf{Interpreting check failures.}
Some requirements depend on others: a missing object cannot follow its requested path.
If a failed check cannot evaluate a requirement because a necessary earlier condition is unmet, we count that dependent requirement as failed.
We mark a mode absent only when all its relevant requirements are confirmed to pass, and leave other inconclusive cases unknown.
This analysis does not change our original verification decisions or benchmark scores.

\textbf{Failure rates and coverage.}
We compute each mode's failure rate only over benchmark instances with a check for the corresponding requirement.
To avoid giving configurations with more attempts extra weight, we calculate each configuration's fraction of attempts with that failure mode on each instance, then average equally across these instances and all 47 configurations.
Coverage is the share of those attempts with enough evidence to decide whether the mode occurred, using the same weights; unknown cases remain in the denominator.
To keep a high rate on a rare requirement from dominating the benchmark-wide ranking, we multiply each rate by the fraction of our 100 instances with a relevant check.
The modes overlap, so these rankings do not describe independent capabilities.
Table~\ref{tab:appendix_failure_modes} reports the failure rates and coverage for all 16 modes.

\begin{table}[t]
\footnotesize
\centering
\setlength{\tabcolsep}{4pt}
\caption{
Failure rates for all 16 modes.
We report the number of benchmark instances with relevant verification checks and the failure rate, with coverage in parentheses (both in percent).
Rows follow the benchmark-wide ranking described in the text.
}
\label{tab:appendix_failure_modes}
\begin{tabular*}{\textwidth}{@{\extracolsep{\fill}}lrc@{}}
\toprule
Failure mode & Instances & Failure rate (coverage) \\
\midrule
Reference-pose mismatch & 98 & 41.50 (94.28) \\
Motion trajectory mismatch & 95 & 38.75 (88.63) \\
Event coordination mismatch & 95 & 32.41 (93.04) \\
Linked-part or joint mismatch & 77 & 39.51 (90.53) \\
Spatial relation or contact mismatch & 51 & 40.47 (94.10) \\
Transform-origin mismatch & 55 & 36.76 (92.45) \\
Easing or speed mismatch & 76 & 23.31 (96.12) \\
Repetition mismatch & 74 & 21.03 (90.55) \\
Shape or scale mismatch & 68 & 21.48 (86.33) \\
Duration or hold mismatch & 51 & 25.29 (88.20) \\
Visibility-state mismatch & 56 & 21.97 (84.00) \\
Occlusion or clipping mismatch & 32 & 27.20 (72.68) \\
Missing or incomplete action & 58 & 14.97 (64.78) \\
Required stationarity mismatch & 57 & 10.33 (94.61) \\
Reveal or erasure mismatch & 8 & 33.78 (99.04) \\
Appearance-preservation mismatch & 14 & 18.47 (92.23) \\
\bottomrule
\end{tabular*}
\end{table}

\subsection{Model Profiles and Qualitative Examples}
\label{app:model_profiles}

\textbf{Near-tied model profiles.}
We identify three pairs whose pass@1 scores in Figure~\ref{fig:model_pass_at_1} are equal or differ by one point.
GPT-6 Astra has more reference-pose mismatch, while Fable 5.1 has more event coordination mismatch.
Grok 4.6 has more reference-pose mismatch, while Muse Spark 1.3 has more shape or scale mismatch.
Gemini 3.8 Flash has more reference-pose mismatch, while GLM 5.3 has more reveal or erasure mismatch.
Table~\ref{tab:appendix_profile_contrasts} gives these selected contrasts.
For these model comparisons, we count a failure only when our verification code and report establish an unmet animation requirement; measurement gaps remain unknown.
We report coverage alongside each rate because a smaller established failure rate need not imply better performance when evidence coverage differs.

\begin{table}[t]
\scriptsize
\centering
\setlength{\tabcolsep}{3pt}
\caption{
Selected failure rates (coverage), in percent, for models with similar pass@1.
Each pair shows two modes that differ in opposite directions; these illustrate the profile differences rather than enumerate every mode.
}
\label{tab:appendix_profile_contrasts}
\begin{tabular*}{\textwidth}{@{\extracolsep{\fill}}lcc@{}}
\toprule
Failure mode & First model & Second model \\
\midrule
\multicolumn{3}{l}{Fable 5.1 (68.0) / GPT-6 Astra (67.0)} \\
Reference-pose mismatch & 3.9 (84.8) & 14.5 (91.4) \\
Event coordination mismatch & 4.3 (85.4) & 2.1 (86.3) \\
\addlinespace
\multicolumn{3}{l}{Grok 4.6 (41.0) / Muse Spark 1.3 (41.0)} \\
Reference-pose mismatch & 27.2 (84.7) & 20.7 (83.6) \\
Shape or scale mismatch & 2.5 (75.7) & 4.1 (78.9) \\
\addlinespace
\multicolumn{3}{l}{Gemini 3.8 Flash (36.0) / GLM 5.3 (36.0)} \\
Reference-pose mismatch & 29.5 (82.9) & 17.1 (82.0) \\
Reveal or erasure mismatch & 6.2 (96.9) & 17.6 (96.5) \\
\bottomrule
\end{tabular*}
\end{table}

\begin{figure}[t]
    \centering
    \includegraphics[width=\textwidth]{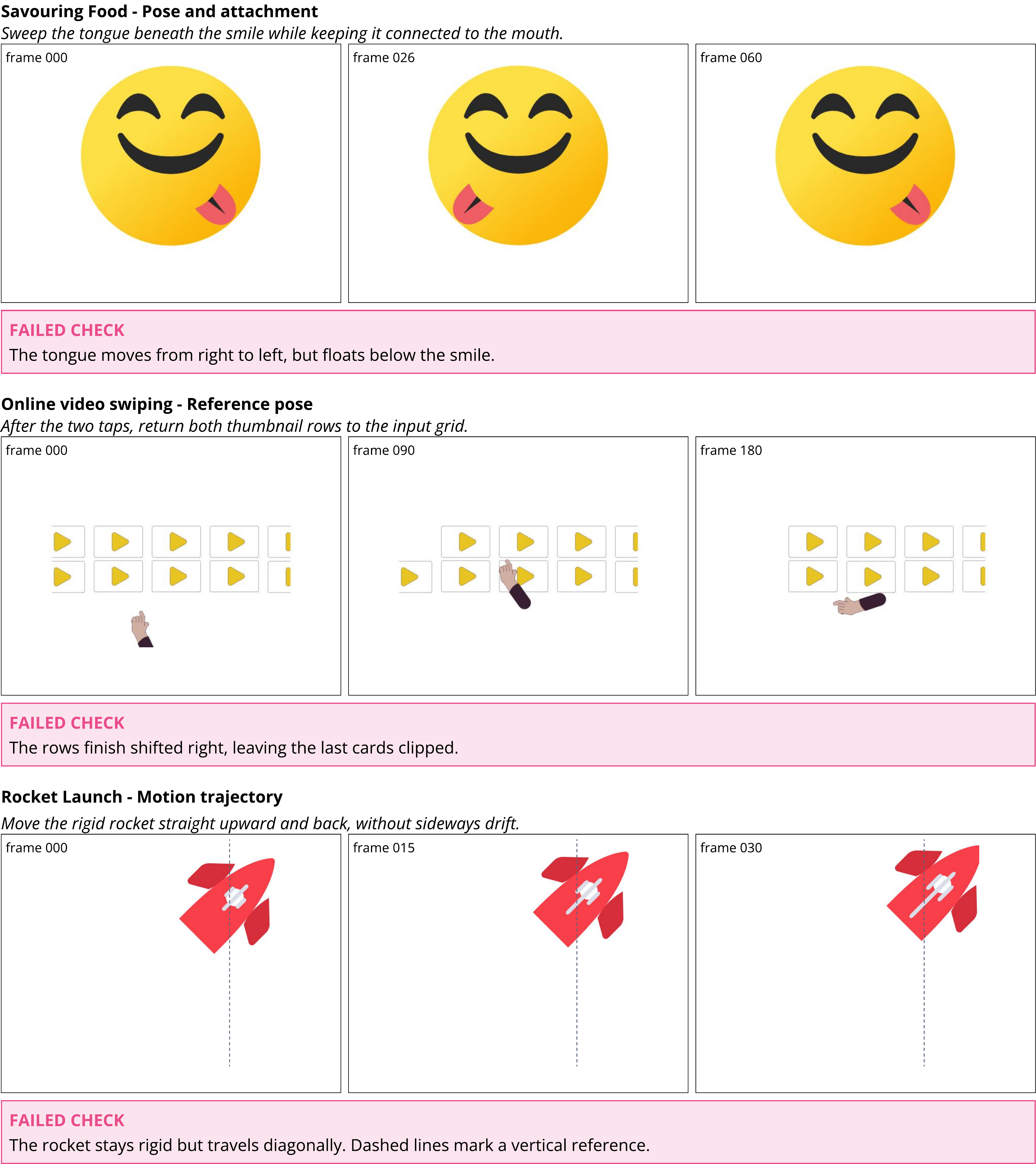}
    \caption{
    \textbf{Additional spatial failure examples.}
    Savouring Food moves its tongue below the smile instead of keeping it connected.
    Online video swiping fails to restore the input grid, leaving cards clipped at the right edge.
    Rocket Launch preserves the rocket assembly but moves it diagonally rather than straight upward.
    Frame labels show the original frame indices.
    }
    \label{fig:appendix_failure_examples}
\end{figure}

\begin{figure}[t]
    \centering
    \includegraphics[width=\textwidth]{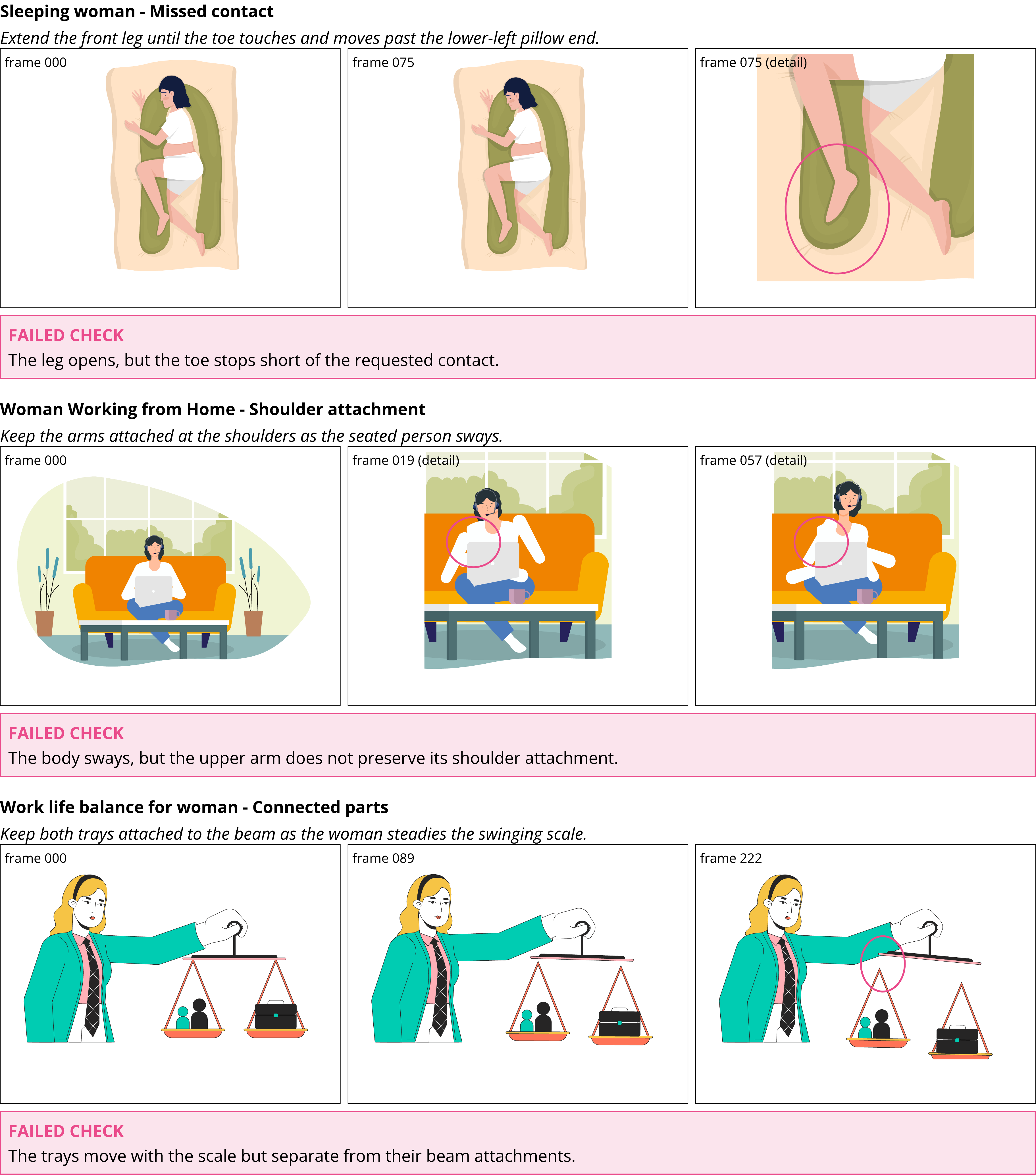}
    \caption{
    \textbf{Human-character failure examples.}
    The sleeping woman's leg extends but her toe stops short of the requested pillow contact.
    Woman Working from Home moves the upper arm away from its shoulder attachment.
    Work life balance for woman lets the scale trays separate from their beam attachments.
    Close-ups expose the missed contact and joint displacement.
    The sleeping-woman example uses poses from our existing human-review collection; all frame labels show the original frame indices.
    }
    \label{fig:appendix_failure_examples_people}
\end{figure}

\begin{figure}[t]
    \centering
    \includegraphics[width=\textwidth]{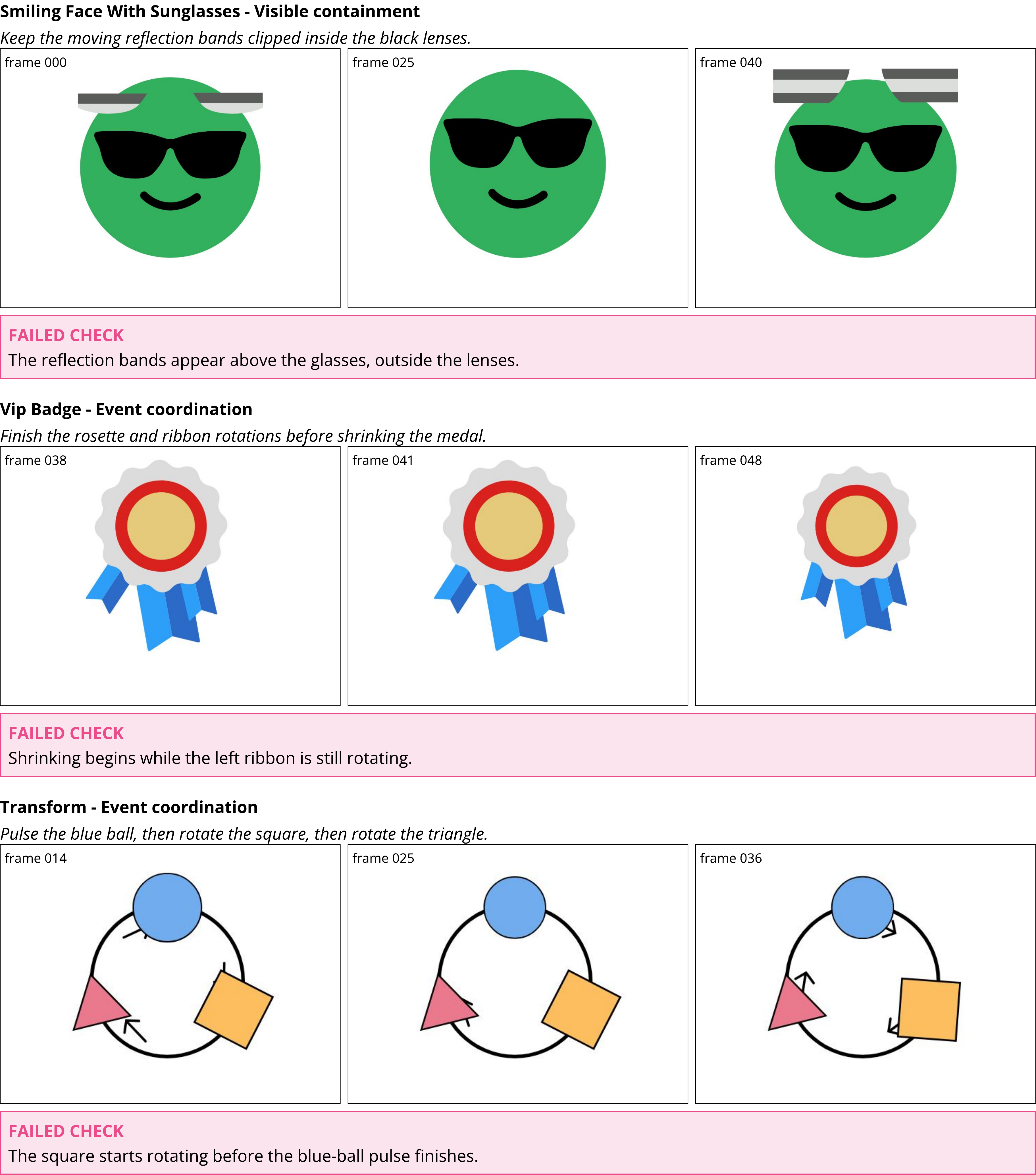}
    \caption{
    \textbf{Additional appearance and timing failure examples.}
    Smiling Face With Sunglasses lets the reflection bands escape the lens boundaries.
    Vip Badge starts shrinking before its ribbon rotation finishes.
    Transform starts the square rotation before the blue-ball pulse ends.
    Each annotation identifies a failure supported by the corresponding saved check report.
    }
    \label{fig:appendix_failure_examples_timing}
\end{figure}

\textbf{Qualitative examples.}
To show how our checks explain failures across different visual designs, Figures~\ref{fig:appendix_failure_examples}--\ref{fig:appendix_failure_examples_timing} add nine examples beyond the three in our main figure.
The examples include human characters, emojis, interface elements, and animated objects.
We show the requested motion, selected generated frames, and a specific failed condition for each example.

\makeatletter
\setlength{\@fptop}{0pt}
\makeatother

